%% file: main.tex
\documentclass{svproc}

\usepackage{url}

\usepackage{stfloats}
\usepackage{longtable}

\usepackage{paralist}
\usepackage{cite}
\usepackage{amsmath,amssymb,amsfonts}
\usepackage{graphicx}
\usepackage{xcolor}

\usepackage{csquotes}

\usepackage{listings}

\begin{document}

\mainmatter              
\title{Fairness Certification for Natural Language Processing and Large Language Models}
\titlerunning{Fairness Certification}  
%
\author{Vincent Freiberger\inst{1} \and Erik Buchmann\inst{2}}
\authorrunning{Freiberger et al.} 
%
\tocauthor{Vincent Freiberger, Erik Buchmann}
\institute{
Dept. of Computer Science, Leipzig University, Germany\inst{1, 2} \\
Center for Scalable Data Analytics and Artificial
Intelligence (ScaDS.AI) Dresden/Leipzig, Germany\inst{1, 2}\\ 
\email{freiberger@cs.uni-leipzig.de}\inst{1}
\email{buchmann@informatik.uni-leipzig.de}\inst{2}}

\maketitle              

\begin{abstract}
Natural Language Processing (NLP) plays an important role in our daily lives, particularly due to the enormous progress of Large Language Models (LLM). However, NLP has many fairness-critical use cases, e.g., as an expert system in recruitment or as an LLM-based tutor in education. Since NLP is based on human language, potentially harmful biases can diffuse into NLP systems and produce unfair results, discriminate against minorities or generate legal issues. Hence, it is important to develop a fairness certification for NLP approaches.

We follow a qualitative research approach towards a fairness certification for NLP. In particular, we have reviewed a large body of literature on algorithmic fairness, and we have conducted semi-structured expert interviews with a wide range of experts from that area. We have systematically devised six fairness criteria for NLP, which can be further refined into 18 sub-categories. Our criteria offer a foundation for operationalizing and testing processes to certify fairness, both from the perspective of the auditor and the audited organization.
\end{abstract}

\input{1-intro}
\input{2-related}
\input{3-approach}

\input{4-evaluation}
\input{5-discussion}

\input{6-conclusion}


\bibliographystyle{IEEEtran}
\bibliography{literature}
\appendix
\input{AppendixInterviewGuide}
\input{AppendixDetailed}

\end{document}

%% file: 1-intro.tex
\section{Introduction}
\label{sec:intro}
Fairness is important for Natural Language Processing (NLP) approaches: 
NLP is used in high-stakes contexts such as healthcare~\cite{wong2018natural}. It is also integrated into daily-use technologies, e.g., voice assistants like Amazon Alexa~\cite{lopatovska2019talk} or AI-based chatbots like ChatGPT~\cite{openai2023chatgpt}.  A lack of fairness often materializes as allocative or representational harm~\cite{blodgett2020language} for marginalized groups~\cite{markl2022language}. An example of allocative harm is a resume filtering system, that prefers male applicants~\cite{sun2019mitigating}. Representational harm would be a translation app that translates to gender stereotypes~\cite{stanovsky2019evaluating}, cultural stereotypes or demeaning language~\cite{caliskan2017semantics, sun2019mitigating, weidinger2022taxonomy}. To avoid harm, a fair NLP application must not only resist gender bias~\cite{bolukbasi2016man,papakyriakopoulos2020bias,tatman2017gender,Ovalle2023}, but also ableist \cite{hassan2021unpacking}, ethnical \cite{bridgeman2012comparison,blodgett2017racial,papakyriakopoulos2020bias}, age-related \cite{diaz2018addressing}, religion-related \cite{brown2020language} or sexuality bias \cite{garg2019counterfactual, papakyriakopoulos2020bias}.
Performant NLP models tend to be opaque and complex ~\cite{danilevsky2020survey, lepri2018fair}. Interactions with such sociotechnical systems are typically also complex \cite{chouldechova2020snapshot}. Hence, it is not practically achievable for users or affected individuals to verify fairness. Fairness certification could be embraced to reduce information asymmetries \cite{cihon2021ai}. 


The concern of this paper is to develop a broad set of criteria, that can be used by an auditor to assess and certify the fairness of an approach that makes use of NLP, be it a large language model, a recruiting-tool or an AI-chatbot for teaching.
This is challenging: 
First, many different definitions of fairness exist~\cite{verma2018fairness,mehrabi2021survey}, and some of them are contradictory~\cite{Chouldechova2017Big, Defrance2023}. 
Second, it is still unclear yet, which fairness criteria are important for NLP approaches and how they impact each other. For instance, residual unfairness may remain after bias mitigation has been resolved~\cite{kallus2018residual}.
Third, efforts towards fair NLP and certifying fairness of related AI approaches exist~\cite{raji2020closing,adler2018auditing,dwork2012fairness,park2022fairness,segal2021fairness, landers2022auditing}. But there is neither an established nor a holistic framework on how fairness could be certified and what could be audited \cite{costanza2022who,petersen2021postprocessing}. 
To approach certification criteria that can be applied in practice, we need to consider the challenges of professionals tasked with auditing the fairness of a system, as well as the challenges of developers that encounter fairness issues when creating NLP systems. Thus, our research question is as follows: 

\textit{What criteria are relevant to consider for fairness certification for NLP approaches from a practitioner's point of view?}


To approach our research question, we strive for broad, qualitative research. We have decided to derive a basic set of auditable fairness criteria from literature on NLP, AI fairness and AI certification. Based on this set of criteria, we have developed a concept for a semi-structured interview with stakeholders from business and research. We have analyzed the interview transcripts in order to find out (a) which measures for ensuring fairness need to be considered for NLP approaches, and (b) how they influence each other. 
We make four contributions: 

\begin{compactitem}
    \item We outline and structure an extensive, up-to-date body of research literature on NLP fairness, AI fairness and fairness auditing from the last years.     %
    \item We describe our qualitative research approach, which is based on a series of semi-structured interviews with 14 experts from various areas related to NLP and algorithmic  fairness.
    \item We devise a hierarchical coding scheme for the certification of fairness for NLP approaches. 
    \item We provide an overview of the six main criteria, with 18 sub-criteria on the second hierarchy level for the auditing of NLP approaches, which we have devised from the interview transcripts. 
\end{compactitem}

To the best of our knowledge, we are the first to devise a holistic, hierarchical coding scheme for the fairness certification of NLP approaches, which is backed up by literature and expert interviews. Our findings allow to develop a fairness certification for a wide range of NLP applications, including large language models and other text-generating AI approaches. Fairness is related to trust \cite{starke2022fairness} and reduces information asymmetries \cite{cihon2021ai}. Thus, our results help to establish a selling point for consumers, particularly marginalized groups. From a legal perspective, a certification can be a precautionary measure \cite{landers2022auditing}. 

\textbf{Paper structure:}
Section~\ref{sec:related} reviews related work and gives a theoretical background. 
Section~\ref{sec:approach} outlines our research method.
The Sections~\ref{sec:eval} and \ref{sec:disc} evaluate and discuss our findings.
Finally, Section~\ref{sec:conclusion} concludes.

%% file: 2-related.tex
\section{Related Work}
\label{sec:related}

In this section, we review existing literature on the Natural Language Processing (NLP) workflow and its modes of use. We provide an overview of fairness in Artificial Intelligence (AI) approaches on which NLP builds. Finally, we describe existing certification approaches for such AI approaches. 

\subsection{Natural Language Processing}
NLP makes human language computable~\cite{kang2020natural}, and can be distinguished into Natural Language Understanding (NLU) and Natural Language Generation (NLG)~\cite{kang2020natural}. NLU refers to a machine’s comprehension of human language and \enquote{extracting valuable information for downstream tasks}~\cite{kang2020natural}. Text summarization~\cite{liddy2001natural,otter2021survey}, intend recognition~\cite{otter2021survey}, machine translation~\cite{otter2021survey}, named entity recognition~\cite{khurana2022natural}, sentiment analysis~\cite{yi2003sentiment} or text classification~\cite{otter2021survey} are examples for NLU. 

NLG refers to producing human-understandable text or speech in natural languages~\cite{mcdonald2010natural}. This is done by predicting the next token in a sequence of words, based on data sources like graphics, video, text, or audio. 
Prominent examples are AI-based Large Language Models (LLMs) like ChatGPT~\cite{openai2023chatgpt}. 

Modern NLP is based on language models that allow the creation of word embeddings~\cite{matthews2021gender,otter2021survey} in the semantic space of a language.
Building an NLP model starts with selecting the corpus, a \enquote{collection of linguistic data, either compiled from written texts or transcribed from recorded speech}~\cite{khurana2022natural}. Models are typically large-scale Deep Learning models, which are pretrained on very large sets of text data~\cite{otter2021survey}. Word embeddings resulting from language models can also be utilized as features for downstream NLP models~\cite{bolukbasi2016man}. Evaluation is typically performed on specific benchmark data sets for the field~\cite{otter2021survey}.

\subsection{Artificial Intelligence Fairness}
Fairness can be defined as the equally performant treatment of all humans by AI, without discriminating against any individuals or communities~\cite{ashok2022ethical}. The whole life cycle of AI needs to be considered, as fairness problems can occur at all stages~\cite{ntoutsi2020bias}. We employ the Cross Industry Standard Process for Data Mining (CRISP-DM)~\cite{wirth2000towards} when referring to the life cycle of AI. 
We chose a broad and generalist definition of fairness to encapsulate different conceptualizations of fairness brought up by interview partners (cf. Sec.~\ref{sec:eval}).
Fairness and its perception are context-dependent and communally derived~\cite{skirpan2017authority,schmidt2017survey}. For measuring fairness, numerous metrics have been discussed in literature~\cite{jacobs2020meaning,verma2018fairness,Defrance2023}, some of which even (partially) contradict each other. 

Assessing fairness is pursued on different levels. This necessitates differentiating between individual fairness and (sub)group fairness, which in themselves have substantial normative differences~\cite{binns2020apparent,mehrabi2021survey}.
Bias~\cite{mehrabi2021survey} is often brought up when fairness is discussed, even if it is not necessarily correlated to fairness~\cite{Cabello2023}. Bias is a \enquote{dynamic and social and not [just] a statistical issue}~\cite{ntoutsi2020bias}. Mitigating biases can be handled during preprocessing, inprocessing or postprocessing~\cite{bellamy2019fairness,friedler2019comparative} in a software life cycle.
Preprocessing captures the data processing before inference. Inprocessing targets the inference process itself. Postprocessing addresses operations performed after inference.

To understand how biases impact fairness, understanding in what ways they are harmful to which specific groups is required~\cite{blodgett2020language}. Biases can result in discrimination, i.e., in differences in the predictive power of models for an individual based on membership in different protected groups~\cite{chen2018why,calmon2017optimized}. It may occur \textit{directly} by utilizing protected attributes such as gender, race, disability, or sexuality. It can also be \textit{indirectly}, via correlations to excluded protected attributes~\cite{mehrabi2021survey}. These attributes may overlap for certain social groups, causing multi-dimensional discrimination like intersectionality~\cite{Roy2023}.  

Different social groups use language differently~\cite{harris2022exploring}. NLP approaches tend to perform worse for marginalized groups because their usage of language is underrepresented~\cite{markl2022language}. Language also transmits beliefs about social groups and imposes labels on them, representing their societal position~\cite{blodgett2020language}. Language use reflects power relationships and social discrimination~\cite{papakyriakopoulos2020bias}. Biases in human language are surfaced in word embeddings~\cite{bolukbasi2016man,papakyriakopoulos2020bias}. 
This is problematic, because word embeddings are used for downstream NLP models, and biases diffuse into these models \cite{papakyriakopoulos2020bias}. There they can cause allocative or representational harm, and must be targeted by debiasing~\cite{bolukbasi2016man,chen2020exploring}.

\subsection{Artificial Intelligence Certification}
According to ISO/IEC 17000:2020~\cite{iso2020conformity}, certification is a \enquote{third party attestation $(\cdots)$ related to an object of conformity assessment}~\cite{iso2020conformity}. The object of conformity assessment refers to an entity to which specific needs and expectations are put into place~\cite{iso2020conformity}. A certification authority performs a provider-independent audit with comprehensive checks to assess conformity with certification criteria, standards, or performance claims~\cite{costanza2022who,ieee2008ieee,lins2022why}. Third-party auditors are \enquote{independent organizations or individuals with no obligation or contractual relationship to the audit target}~\cite{costanza2022who}. 
A certification communicates information, signals quality to the system user, and assurance raises trust~\cite{lins2022why,lansing2018unblackboxing}. This is important in high-stakes domains \cite{20.500.11850/621931}. It helps providers to show their legitimacy and to improve their services and products~\cite{landers2022auditing}. Certifications are defined by content, source, and process elements~\cite{lansing2018unblackboxing, 20.500.11850/621931}. Content captures the specific subject of assessments in an audit. The source describes the institution providing accreditation. Process describes how an assessment for certification is conducted.

For AI auditing, use-case-specific approaches have been suggested~\cite{costanza2022who}. 
Common auditing standards do not exist~\cite{costanza2022who,raji2020closing} yet. This is problematic, as such systems are increasingly deployed in high-stakes domains~\cite{raji2020closing}. Current fairness audits~\cite{adler2018auditing,park2022fairness,segal2021fairness} focus on quantitative aspects. Such audits evaluate system outputs according to mathematical fairness definitions without procedural or qualitative assessment, and without considering intersectionality~\cite{costanza2022who}. This does not necessarily result in a fair system.

%% file: 3-approach.tex
\section{Our Research Method}
\label{sec:approach}


Because fairness certification of NLP approaches~\cite{costanza2022who,myers2020qualitative} is a novel, barely defined research topic (cf.~Sec.~\ref{sec:related}), we approach our research question with semi-structured expert interviews. Such interviews bring consistency, focus, and structure to the interview while offering room for improvisation~\cite{corbin2015basics,myers2020qualitative}. 
In particular, we follow Myers' qualitative research methodology~\cite{myers2020qualitative} and Corbin \& Strauss' approach to data analysis~\cite{corbin1990grounded,corbin2015basics}. 
%

%
Our research method consist of the steps \textit{literature review}, \textit{interview guide}, \textit{interviewee selection}, \textit{conducting} and \textit{transcribing} the interviews, \textit{coding} the interview transcripts, \textit{annotation and reflection} and \textit{refinement} of the interviews. 
In the following, we explain these 8 steps. 

\textbf{Literature Review:} This step provides us with all the information necessary to develop an interview guide and to interpret the interview results. In particular, we need an overview of NLP, AI fairness, and certification (cf. Sec.~\ref{sec:related}). 

\textbf{Interview Guide:}
On the basis of the literature review, we developed an interview guide with open questions (see Appendix \ref{tab:guide}). 
Our guide structures the interview into 8~parts. 
In the first part, (1) we welcome our interviewees and (2) explain the organization of the interview. 
Next, (3) we capture the professional background of the interviewee, and (4) clarify the terminology for the interview. 
Then (5) we ask for criteria for NLP development and (6) for sustaining NLP fairness over time. 
After that, (7) we leave room for open topics to talk about.
Finally, (8) we thank our interviewee and conclude the interview.

\textbf{Interviewee Selection:}
Our research question calls for practically applicable research results drawn from the entire life cycle of NLP approaches. 
To investigate relevant criteria for fairness certification, we decided to select interviewees as follows: 
The interviewees should be working in the private sector on NLP in a relevant role like data scientist, consultant, or manager. They should have a minimum of two years of industry experience in NLP and, ideally, exposure to fairness in AI. To broaden our findings \cite{jakesch2022different,myers2020qualitative}, we wanted to gain diverse experts in the dimensions of age, gender, and cultural background.

To identify matching candidates, we browsed LinkedIn and Xing as well as the authors' professional networks. Matching candidates received a flyer with basic information on the research project and the interview guide in advance, to let them judge whether they could provide value and reflect on the topic. 

On this basis, we selected 12 interviewees. In addition, we made an exception to include two more interviewees with a background in academia.
These interviewees were involved in algorithmic audits. Both had more than three years of experience in algorithmic fairness. 
Table \ref{table:1} gives an overview of the interview partners in the order in which they were interviewed. The first column of the table contains the ID of the interview partner. In the following, we will use the ID to relate a statement to a person. The second and third columns describe the role of the interviewee and the business area of the interviewee's company. The last column contains the interviewee's professional expertise.

\newcommand{\intvwr}[1]{\textit{I$_{#1}$}}
\begin{table}[ht]
\centering
\caption{Overview of interview partners}
\begin{tabular}{c|p{2.8 cm}|p{3.5 cm}|p{4.2 cm}}
\textbf{ID}	& \textbf{Profession}	& \textbf{Affiliation}	& \textbf{Expertise}\\
\hline
\intvwr{1}	& CTO	& Regional, \par AI Solutions	& NLU, whole life cycle\\
\hline
\intvwr{2}	& Data Scientist	& International, \par IT Consulting	& Wide variety in NLP, whole life cycle\\
\hline
\intvwr{3}	& Data Scientist	& National, \par IT Services	& Wide variety in NLP, whole life cycle\\
\hline
\intvwr{4}	& Team Lead NLP	& National, \par AI and Research	& Wide variety in NLP\\
\hline
\intvwr{5}	& CTO	& National, Conversational AI Solutions	& NLG\\
\hline
\intvwr{6}	& Head of NLP	& National, \par IT Services	& NLG\\
\hline
\intvwr{7}	& Data Scientist	& International, \par IT	& NLU, search query handling\\
\hline
\intvwr{8}	& Associate Vice \par President AI	& International, \par IT Consulting	& Wide variety in NLP\\
\hline
\intvwr{9}	& Senior Program \par  Manager	& International, IT	& Conversational AI, ASR\\
\hline
\intvwr{10}	& Consultant	& Multinational, \par IT Consulting	& NLU	\\
\hline
\intvwr{11}	& Tech Lead NLP	& International, \par IT Consulting	& Wide variety in NLP\\
\hline
\intvwr{12}	& Research Scientist	& National, Research	& Algorithmic fairness\\
\hline
\intvwr{13} & Co-Founder	& National, \par IT Services	& NLU, focus: preprocessing\\
\hline
\intvwr{14}	& Senior Applied \par Scientist	& Multinational,\par  E-Commerce	& Algorithmic fairness, \par certification\\
\end{tabular}
\label{table:1}
\end{table}

\textbf{Conducting the Interviews:}
We conducted one interview in English and 13 in German, all of them via video conferences. 
The interviews followed the interview guide, as explained in the second step. 
During the main section of the interview, we generally encouraged further input on topics verbally and nonverbally by active listening, open questions, and reflecting back to the interviewee \cite{myers2020qualitative}. 
The average duration of all 14 interviews was 51 minutes.

\textbf{Transcribing the Interviews:}
We transcribed all interviews manually, following the guidelines of Dresing and Pehl \cite{dresing2018praxisbuch}. 
We adapted the guidelines by leaving out timestamps, which were not needed. To ensure correctness, we asked the interviewees to check our transcripts.

\textbf{Coding the Transcriptions:}
To gain insight from the transcripts, we developed a coding scheme in an iterative approach based on open and axial coding~\cite{corbin1990grounded}. 
Open coding allows us to freely name concepts represented in interview data. We use axial coding to understand concepts’ context, cause, and consequence. We identify concepts for the coding scheme by interviewees naming them explicitly, by abstracting from specific ideas to concepts, or by identifying examples for a concept. 

\begin{figure}[ht]
   \centering
   \includegraphics[width=0.85\linewidth]{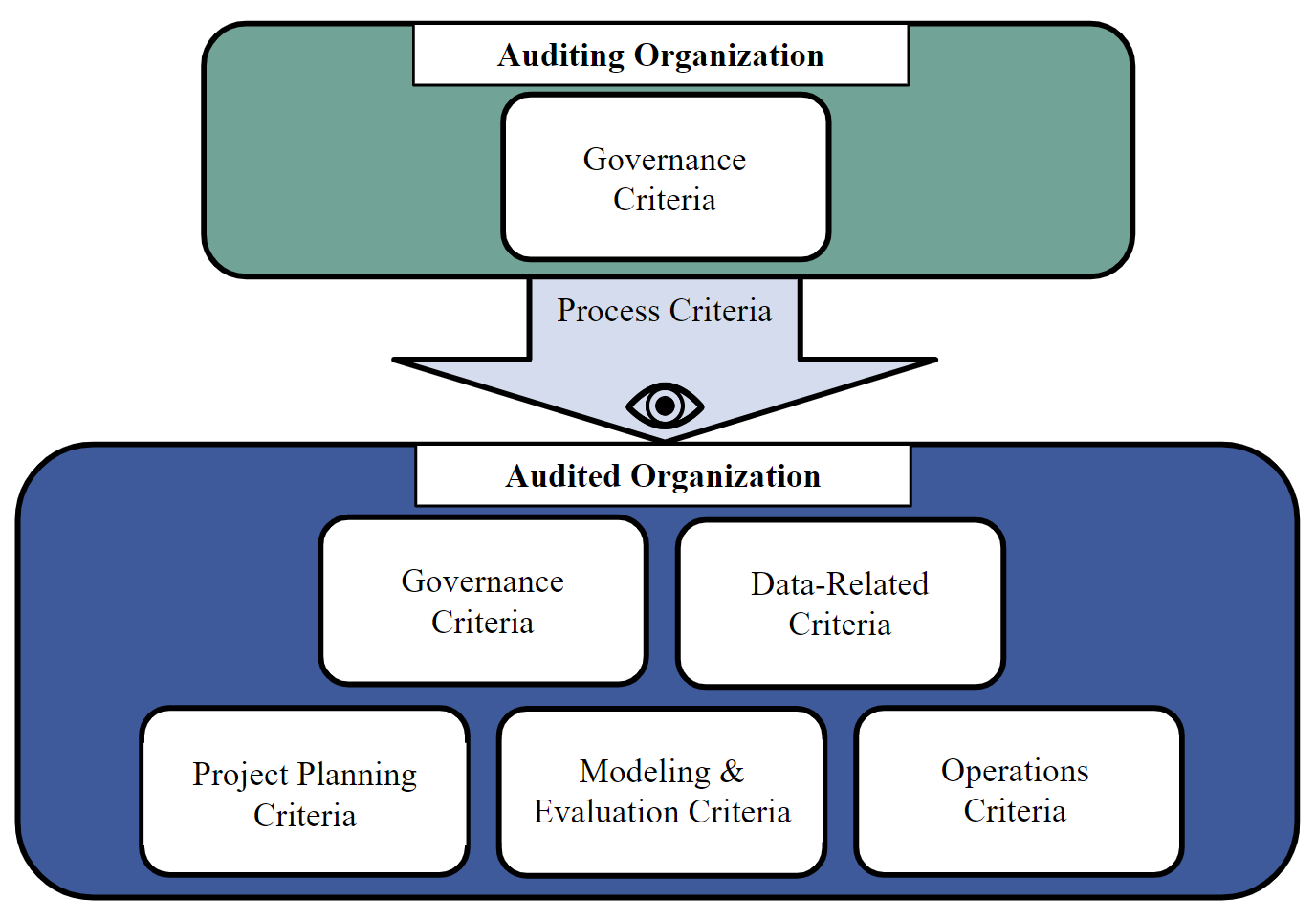}   
   \caption{Top-level codes for the fairness certification of NLP approaches}
   \label{fig:1}
\end{figure}

In advance of our results in Sec.~\ref{sec:eval}, Figure~\ref{fig:1} shows the main criteria identified in our coding scheme. On this level, the coding scheme consists of six criteria, that are relevant for NLP certification. 
The scheme is divided into the auditing organization and the audited one. \textbf{Governance Criteria} are relevant for the certification process for both of them, from opposite points of view. The general characteristics and aims of the certification are explained by \textbf{Process Criteria}. Finally, \textbf{Data-Related Criteria}, \textbf{Project Planning Criteria}, \textbf{Modeling \& Evaluation Criteria} and \textbf{Operations Criteria} specify processes and contents of the audited organization. Note that coding is an iterative process, i.e., Figure~\ref{fig:1} illustrates the last version of dynamic refinement.

\textbf{Annotation and Reflection:}
We annotated 1162 text passages, utilizing memos to capture the central ideas and concepts represented in the passage. After multiple iterations over the scheme, 1095 coded passages contained 587 overall codes, with 124 belonging to open coding and 473 to axial coding.

\textbf{Refinement:}
Conducting interviews is a dynamic process. After conducting and analyzing our first four interviews, we adapted our interview guide. 
In particular, we worked on our questions to simplify the interview process and to avoid unnecessary follow-up questions.
Furthermore, we learned that our broad approach puts a lot of time pressure on the interviewer, which does not allow us to ask in-depth questions. 
Thus, we decided to ask our interviewees about prior experiences and thoughts on fairness certification first. This allows us to skip topics in the subsequent interview, that the interviewees did not feel qualified enough to talk about.

%% file: 4-evaluation.tex
\section{Interview Findings}
\label{sec:eval}

\begin{figure}[ht]
   \centering
   \includegraphics[width=1\textwidth,trim=0 0 0 7px, clip]{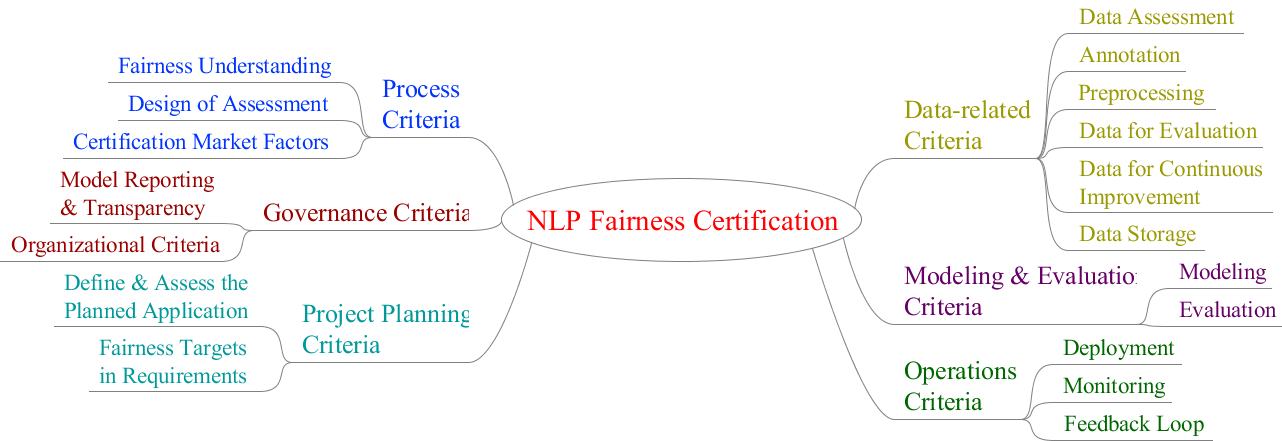}   
   \caption{Mind map of the coding scheme for the fairness certification of NLP approaches}
   \label{fig:2}
\end{figure}

This section gives an overview of all criteria for NLP certification that we have identified. Figure~\ref{fig:2} illustrates our hierarchy of codes and sub-codes. Note that we further subdivided the sub-codes, which we omit here for lack of space. Appendix~\ref{sec:appendixeval} contains a more detailed description of our results. 

In the following, we explain the coding of our six criteria together with their sub-criteria captured by sub-codes. The intervieweees who addressed them are denoted in brackets. Criteria that were mentioned by many interviewees tend to be more important for a fairness certification of NLP approaches.

\textbf{Process Criteria} refer to the implementation of a fairness certification process. Our interviewees identified three distinguished aspects of process criteria, which we modeled as sub-codes. 

The sub-code \textit{Fairness Understanding} captures properties of fairness conception, that should be considered for fairness certification according to our interviews. Examples include cultural dependence (\intvwr{9}, \intvwr{13}, \intvwr{14}), use case dependence (\intvwr{1}, \intvwr{3}, \intvwr{5}, \intvwr{6}, \intvwr{7}, \intvwr{8}, \intvwr{12}, \intvwr{13}, \intvwr{14}), dynamic (\intvwr{13}, \intvwr{14}) and non-binary (\intvwr{12}) properties. It also covers the interviewees’ conceptions of what is to be considered fair, which is nuanced, and varies between conceptions like conditional statistical parity (\intvwr{1}, \intvwr{4}, \intvwr{6}, \intvwr{7}), counterfactual fairness (\intvwr{3}, \intvwr{4}, \intvwr{5}, \intvwr{10}, \intvwr{13}) and notions of inclusiveness (\intvwr{2}, \intvwr{4}, \intvwr{7}, \intvwr{8}, \intvwr{9}, \intvwr{11}). 

The sub-code \textit{Certification Market Factors} covers how a certification and regulation may affect each other (\intvwr{1}, \intvwr{3}, \intvwr{4}, \intvwr{6}, \intvwr{14}), the market which may build around certification (\intvwr{1}) and the adoption of certification dependent on alignment with corporate goals (\intvwr{1}, \intvwr{10}, \intvwr{11}, \intvwr{13}, \intvwr{14}). 

In \textit{Design of Assessment}, we capture properties for an assessment and its scope. A requirement is finding a way to create an audit process that is as holistic as possible (\intvwr{1}, \intvwr{6}, \intvwr{8}, \intvwr{9}, \intvwr{11}, \intvwr{12}, \intvwr{13}, \intvwr{14}). A certification shouldn't put small companies at a disadvantage (\intvwr{3}, \intvwr{5}, \intvwr{6}, \intvwr{11}). Interviewees discuss the approach to certification. They propose, for instance, to certify individuals instead of processes (\intvwr{5}, \intvwr{6}, \intvwr{8}). Building a certification on existing standards and best practices is also agreed upon (\intvwr{1}, \intvwr{5}, \intvwr{8}, \intvwr{10}, \intvwr{11}, \intvwr{13}, \intvwr{14}). Another major point of discussion was to what extent a certification process should be risk-dependent in its scope (\intvwr{1}, \intvwr{2}, \intvwr{9}, \intvwr{14}) or in its obligation to be performed (\intvwr{2}, \intvwr{10}, \intvwr{11}).

\textbf{Governance Criteria} refer to all measures the certified organization should undertake to ensure the integrity of its processes and management. 
Recall that governance is important both from the perspective of the auditing and the audited organization. 
We have identified two sub-codes for governance criteria, which encompass several aspects. 

The sub-code \textit{Model Reporting \& Transparency} follows the idea of disclosing biases, basic information, data understanding, model explanations, and evaluation. 
For basic information, our interviewees want fields of application (\intvwr{4}, \intvwr{13}), the underlying fairness definition, and the frameworks the system was built on (\intvwr{4}) to be disclosed. 
Data understanding covers all relevant information about the data the system utilizes and how it is processed (\intvwr{4}, \intvwr{12}, \intvwr{13}). 
Model explanations cover disclosing the model with its architecture (\intvwr{4}, \intvwr{8}, \intvwr{12}) and explainability of the model (\intvwr{2}, \intvwr{3}, \intvwr{4}, \intvwr{6}, \intvwr{8}, \intvwr{12}). 
For evaluation, interviewees bring up that reporting should contain what has been tested regarding the robustness of the model against biases (\intvwr{2}, \intvwr{4}). Evaluation metrics and their resulting data from testing should be publicly documented (\intvwr{4}, \intvwr{8}). 
Reporting data bias (\intvwr{4}), model bias (\intvwr{3}), and countermeasures against biases that are implemented (\intvwr{3}, \intvwr{13}) should also be done.
As disclosure mechanisms, our interviewees suggest utilizing Model Cards (\intvwr{2}, \intvwr{3}, \intvwr{4}, \intvwr{8}, \intvwr{12}, \intvwr{13}, \intvwr{14}), open sourcing the model entirely (\intvwr{8}) or giving the users an option to access relevant information to the prediction the model made for them on demand (\intvwr{8}).

The sub-code \textit{Organizational Criteria} targets both the audited and the auditing organization. 
Interviewees highlight that the context of the auditing organization should be reflected because it can impede its ability to make neutral and holistic assessments. This involves the region it is located in (\intvwr{13}), the political system it is integrated into (\intvwr{13}), its initiator (\intvwr{13}), its integration in the economic system (\intvwr{2}, \intvwr{10}), or the lack in diversity of professional backgrounds of its auditors (\intvwr{6}, \intvwr{9}). The auditing organization consolidates best practices (\intvwr{2}, \intvwr{6}, \intvwr{9}, \intvwr{12}).
For the audited organization the diversity of the development team (\intvwr{2}, \intvwr{4}, \intvwr{10}, \intvwr{11}, \intvwr{13}), internal accountability mechanisms and employee qualification and training should be checked. 
Internal accountability involves checking for internal assessments regarding fairness that are put in place by the audited organization (\intvwr{7}, \intvwr{14}). Roles should be assigned clear responsibilities regarding fairness (\intvwr{1}, \intvwr{2}, \intvwr{3}, \intvwr{6}, \intvwr{7}, \intvwr{9}, \intvwr{10}, \intvwr{11}, \intvwr{14}). Interviewees mention that accountability also needs someone who is responsible for fairness in the organization and supervision targeted at fairness (\intvwr{3}, \intvwr{5}, \intvwr{6}, \intvwr{7}, \intvwr{11}).
Employee qualification \& training are seen by interviewees to provide a reasonable job fit for employees (\intvwr{3}, \intvwr{6}). A holistic overview of the process of training and operating an NLP system and what could go wrong regarding fairness (\intvwr{6}, \intvwr{8}) should also be provided. Moreover, the interviews reveal the importance of raising general awareness of bias issues and ensuring general knowledge about biases (\intvwr{1}, \intvwr{3}, \intvwr{4}, \intvwr{6}, \intvwr{7}, \intvwr{11}).

\textbf{Project Planning Criteria} address processes in the planning stage of a product and its underlying business problem that the certification institution should assess. Their scope ends right before data understanding in the CRISP-DM framework. This involves two sub-codes. 

For sub-code \textit{Define \& Assess the Planned Application}, the interviewees agree about checking definitions of use case (\intvwr{4}, \intvwr{8}, \intvwr{9}, \intvwr{12}, \intvwr{13}, \intvwr{14}) and stakeholders (\intvwr{6}, \intvwr{7}, \intvwr{8}, \intvwr{11}) provided by the audited organization. This entails an assessment for the latter regarding fairness issues or vulnerabilities to fairness risks. This involves checking that mechanisms are in place to properly understand users and stakeholders, as well as the fairness properties of the use case itself. Moreover, the planned solution should fit the use case, also from a fairness perspective (\intvwr{3}, \intvwr{4}, \intvwr{5}, \intvwr{12}). 
The sub-code \textit{Fairness Targets in Requirements} implies checking if a standardized set of requirements has been established to ensure fairness regarding the project over its life cycle (\intvwr{2}, \intvwr{4}, \intvwr{9}).

With \textbf{Data-related Criteria}, we refer to all assessment processes regarding data used for training and retraining as well as regarding procedures for handling and transforming data. Those procedures are described by six sub-codes. 

The sub-code \textit{Data Assessment} comprises an assessment of data quality regarding fairness by checking representativeness via distribution checks (\intvwr{1}, \intvwr{4}, \intvwr{8}, \intvwr{9}, \intvwr{13}) and by assessing regarding harmful, toxic or incorrect data (\intvwr{2}, \intvwr{4}, \intvwr{10}, \intvwr{13}, \intvwr{14}). Interviewees suggest introducing processes for bias checks (\intvwr{1}, \intvwr{2}, \intvwr{4}, \intvwr{5}, \intvwr{7}, \intvwr{9}, \intvwr{11}) and assessments of the sourcing (\intvwr{1}, \intvwr{2}, \intvwr{4}, \intvwr{5}, \intvwr{6}, \intvwr{8}, \intvwr{12}, \intvwr{13}) and collection (\intvwr{3}, \intvwr{12}, \intvwr{14}) of data.  
Sub-code \textit{Annotation} captures characteristics of the annotation process to support fairness. Annotating fairness relevant attributes (\intvwr{1}, \intvwr{2}, \intvwr{11}) is suggested by interviewees. Moreover, comprehensive annotation guidelines (\intvwr{4}, \intvwr{6}, \intvwr{7}) and inter-annotator agreements (\intvwr{4}, \intvwr{7}, \intvwr{10}, \intvwr{13}) are important considerations. 
Sub-code \textit{Preprocessing} focuses on requirements for filtering (\intvwr{3}, \intvwr{4}, \intvwr{13}), selecting (\intvwr{1}, \intvwr{4}, \intvwr{7}, \intvwr{8}, \intvwr{10}, \intvwr{12}, \intvwr{13}), anonymizing (\intvwr{3}, \intvwr{5}, \intvwr{7}, \intvwr{14}) or mapping the data robustly (\intvwr{2}) to ensure fairness. 
The sub-code \textit{Data for Evaluation} entails that suitable, representative data is used (\intvwr{2}, \intvwr{9}, \intvwr{11}) and discusses where to take it from. Interviewees highlight the importance of fairness invariance testing (\intvwr{2}, \intvwr{3}). 
Sub-code \textit{Data for Continuous Improvement} comprises a monitoring process during the operation of the model and a feedback loop targeting improvements and counteracting fairness issues. As criteria for monitoring, interviewees mention fairness test sets (\intvwr{2}, \intvwr{3}, \intvwr{6}, \intvwr{8}, \intvwr{9}), drift monitoring (\intvwr{2}, \intvwr{3}, \intvwr{4}, \intvwr{5}, \intvwr{6}, \intvwr{7}, \intvwr{13}), as well as a request assessment for underrepresented groups (\intvwr{2}) as practices to be implemented by the audited company. Feedback loop practices regarding data need to consider how in-use data like user behavior, usage data or prediction confidence (\intvwr{5}, \intvwr{9}, \intvwr{10}) and user-made corrections (\intvwr{4}) are handled to improve or maintain fairness.
Finally, the sub-code \textit{Data Storage} covers how data should be stored in development and operations (\intvwr{10}) and cached in operations (\intvwr{1}).


The \textbf{Modeling \& Evaluation Criteria} include the two sub-codes Modeling and Evaluation, which can be explained as follows: 

\textit{Modeling} is about embedding fairness into the model architecture itself. Interviewees mention integrating hard-coded elements into the system to ensure fairness (\intvwr{4}, \intvwr{5}, \intvwr{6}, \intvwr{14}), training a separate fairness classifier for the model outputs (\intvwr{5}), working with constraints (\intvwr{2}) or embedding fairness into the model's optimization itself (\intvwr{4}, \intvwr{5}, \intvwr{7}, \intvwr{10}). 

\textit{Evaluation} investigates what tests should be conducted to evaluate a model and how to assess the fairness of a trained model. 
At the center of this interviewees discuss functional testing criteria which take an outside perspective on the system’s outputs given some specific inputs. Interview partners came up with six concepts relevant to this which are validation of predictions (\intvwr{1}, \intvwr{4}, \intvwr{6}), involving affected stakeholders in evaluation (\intvwr{5}), adversarial testing (\intvwr{1}, \intvwr{4}, \intvwr{7}), data sets or benchmarks for testing (\intvwr{2}, \intvwr{3}, \intvwr{6}, \intvwr{8}, \intvwr{9}, \intvwr{11}, \intvwr{13}), criteria focused on human-computer interaction (\intvwr{2}, \intvwr{7}, \intvwr{11}) and ethics criteria (\intvwr{4}, \intvwr{11}).
Another important consideration for evaluation is what metrics should be utilized to test a system's fairness. Interviewees name metrics focused on robustness or generalization as an essential element to measure the model's likelihood of generating unexpected, unfair results (\intvwr{2}, \intvwr{4}, \intvwr{8}). Impact-based metrics depend on the chosen fairness paradigm (\intvwr{3}, \intvwr{7}, \intvwr{8}, \intvwr{12}, \intvwr{14}). Another approach proposed by \intvwr{10} introduces a metric that penalizes the model's use of sensitive attributes or unethical content when calculating loss.

\textbf{Operations Criteria} center around what should be considered regarding fairness when deploying a model and what mechanisms must be implemented to maintain fairness over time. 
That subsumes three sub-codes. 

Sub-code \textit{Deployment} centers around assessments suited to shipping different model sizes and pruning models, which affect their fairness behavior (\intvwr{2}, \intvwr{3}, \intvwr{6}). 
Interviewees define the sub-code \textit{Monitoring} as a continuous assessment procedure of the audited company. This is to ensure its system’s fairness over time, involving fairness tests with the roll-out of updates (\intvwr{1}, \intvwr{2}, \intvwr{3}, \intvwr{7}) and intervention strategies to counteract adversarial or unintentional, but harmful, misuse (\intvwr{9}, \intvwr{10}, \intvwr{13}). 
Finally, with the sub-code \textit{Feedback Loop}, interviewees aim to get the flow of information back from the operational model assessed. This involves checking for the option of flagging system outputs that are perceived as unfair by users and for an option for users to make corrections (\intvwr{1}, \intvwr{2}, \intvwr{3}, \intvwr{7}). Interviewees consider leveraging the data which is acquired by that to understand fairness issues (\intvwr{1}, \intvwr{2}, \intvwr{3}, \intvwr{7}) and, in some cases, trigger retraining (\intvwr{1}, \intvwr{3}, \intvwr{4}). An assessment includes checking if data utilized for retraining the system in continuous improvement is subjected to the same assessments, filtering and preprocessing steps as the initial data (\intvwr{10}).


Table~\ref{tab:fullwidth} provides an overview of which of our interviewees addressed which (sub-)code.
A comparison of this table with Table~\ref{table:1} indicates that we indeed obtained a broad range of experts, which are likely to cover the entire range of the fairness certification process for NLP approaches. 

\begin{table}[ht]
  \centering
  \caption{Second-level coding scheme for fairness certification of NLP approaches}
  \setlength{\tabcolsep}{1.5pt} 
    \begin{tabular}{|p{2.5 cm}|p{3 cm}|*{14}{c|}}
    \hline 
            && \textbf{\intvwr{1}} & \textbf{\intvwr{2}} & \textbf{\intvwr{3}} & \textbf{\intvwr{4}} & \textbf{\intvwr{5}} & \textbf{\intvwr{6}} & \textbf{\intvwr{7}} & \textbf{\intvwr{8}} & \textbf{\intvwr{9}} & \textbf{\intvwr{10}} & \textbf{\intvwr{11}} & \textbf{\intvwr{12}} & \textbf{\intvwr{13}} & \textbf{\intvwr{14}} \\
    \hline
    \textbf{Process} \par \textbf{Criteria} & Fairness \par Understanding &\checkmark &\checkmark &\checkmark &\checkmark &\checkmark &\checkmark &\checkmark &\checkmark &\checkmark &\checkmark &\checkmark &\checkmark &\checkmark &\checkmark\\   
    \cline{2-16}  
    & Design of Assessment &\checkmark &\checkmark &\checkmark &\checkmark &\checkmark &\checkmark & &\checkmark &\checkmark &\checkmark &\checkmark &\checkmark &\checkmark&\checkmark \\
     \cline{2-16}   
     & Certification Market Factors &\checkmark & & & & & & & & & & & &\checkmark& \\
     \hline
     \textbf{Project} \par \textbf{Planning} \par \textbf{Criteria} & Define \& Assess the Planned Application &&\checkmark & &\checkmark &\checkmark &\checkmark &\checkmark &\checkmark &\checkmark &\checkmark &\checkmark &\checkmark &\checkmark &\checkmark \\
     \cline{2-16}
     & Fairness Targets in Requirements & &\checkmark & &\checkmark & & & & &\checkmark & & & & &\\
    \hline
      \textbf{Data-related Criteria} & Data Assessment &\checkmark &\checkmark &\checkmark &\checkmark &\checkmark &\checkmark &\checkmark &\checkmark &\checkmark &\checkmark &\checkmark &\checkmark &\checkmark&\checkmark \\  
    \cline{2-16}
    & Annotation&\checkmark &\checkmark & &\checkmark &\checkmark &\checkmark &\checkmark & & &\checkmark &\checkmark &\checkmark &\checkmark& \\ 
    \cline{2-16}
     & Preprocessing  &\checkmark & &\checkmark &\checkmark &\checkmark & &\checkmark &\checkmark & &\checkmark &\checkmark &\checkmark &\checkmark &\checkmark\\
    \cline{2-16}
    & Data for Evaluation& &\checkmark &\checkmark & & &\checkmark &\checkmark &\checkmark & & & &\checkmark &\checkmark& \\
    \cline{2-16}  
    & Data for Continuous Improvement & &\checkmark &\checkmark &\checkmark &\checkmark &\checkmark &\checkmark & &\checkmark &\checkmark & & &\checkmark& \\
    \cline{2-16}
    & Data Storage &\checkmark & & & & & & & & &\checkmark & & && \\ 
    \hline
    \textbf{Modeling \&} \par \textbf{Evaluation}  \par \textbf{Criteria} & Modeling & &\checkmark &\checkmark &\checkmark &\checkmark &\checkmark &\checkmark & & &\checkmark & & &&\checkmark \\
    \cline{2-16}
    & Evaluation &\checkmark &\checkmark &\checkmark &\checkmark &\checkmark &\checkmark &\checkmark &\checkmark &\checkmark &\checkmark &\checkmark &\checkmark &\checkmark&\checkmark \\
    \hline
    \textbf{Operations Criteria}  & Deployment & &\checkmark&\checkmark & & &\checkmark & & & & & & && \\ 
    \cline{2-16}
    & Monitoring &\checkmark &\checkmark &\checkmark & &\checkmark & &\checkmark &\checkmark &\checkmark & & & &\checkmark& \\  
    \cline{2-16}
     & Feedback Loop &\checkmark &\checkmark &\checkmark &\checkmark & &\checkmark &\checkmark &\checkmark &\checkmark &\checkmark & & && \\
    \hline
    \textbf{Governance Criteria} & Model Reporting \& Transparency &\checkmark &\checkmark &\checkmark &\checkmark & &\checkmark &\checkmark &\checkmark & & &\checkmark &\checkmark &\checkmark& \\ 
    \cline{2-16}
    & Organizational \par Criteria  &\checkmark &\checkmark &\checkmark &\checkmark & &\checkmark &\checkmark &\checkmark &\checkmark &\checkmark &\checkmark &\checkmark&\checkmark &\checkmark \\
    \hline 
\end{tabular}%
\label{tab:fullwidth}%
\end{table}%

%% file: 5-discussion.tex
\section{Discussion}
\label{sec:disc}

In this section, we briefly discuss our results. 
We were interested to learn the entire range of criteria that must be considered to establish fairness certification for an NLP approach. To this end, we have conducted and coded 14 semi-structured interviews with a wide span of diverse experts from business and research. Thus, we think that our findings are well applicable to certifying fairness in a corporate environment. However, we did not cover other sectors, such as public or military. Furthermore, we might not have reached theoretical saturation with 14 interviewees. Moreover, designing and testing a fairness certification process itself was beyond our scope.

One might wonder if this work on fairness and bias was influenced by bias itself. We explicitly tried to exclude the following biases: Selection bias (selecting interview partners on personal preferences), bias in materials (providing documents before the interviews could have influenced the interviewees), verbal/nonverbal bias (due to misunderstandings between interviewer and interviewee) and bias in data analysis (subjective coding). 

%

We observed, that our interviewees focused particularly on criteria for data and functional testing of solutions. 
However, there is a bias in the relevance perception of modeling between NLP fairness research and interviewees' perspectives. 
Existing research already considers model architectures inhibiting social biases, for instance, via regularization \cite{kamishima2012fairness,veale2017fairer,yurochkin2020sensei}, adversarial training \cite{sun2019mitigating,yurochkin2020sensei} or adapting the loss function to support fairness \cite{dwork2018decoupled,kamiran2012decision}. Even though approaches mentioned in interviews are consistent with the literature, the topic was barely mentioned or deemed relatively unimportant in interviews (\intvwr{1}, \intvwr{5}). Future research may investigate this mismatch.


%% file: 6-conclusion.tex
\section{Conclusion}
\label{sec:conclusion}
Fairness certification for natural language processing approaches such as large language models, AI-based chatbots or healthcare applications is an important issue, that is still unresolved.
In this paper, we have conducted and analyzed 14 semi-structured expert interviews with mostly NLP experts in the industry and two algorithmic fairness experts in academia. Our interviewees helped us identify six main criteria and 18 criteria on the second hierarchy level of an open coding scheme for certifying the fairness of an NLP approach. 
Those criteria are an important building block towards operationalizing and testing NLP processes to certify fairness, from the perspective of the auditor as well as from the perspective of the audited organization.

Our interviewees have raised plenty of open questions for future research. For instance: How should a certification process handle the use case dependence of fairness or its non-binary nature and subjectiveness while in a dynamic environment? To what extent would it make sense to make such a certification mandatory? On which best practices and standards should a certification be built? How extensive should it be?
Finally, how must a certification process be structured to specifically address large language models?

%% file: AppendixInterviewGuide.tex
\section{Interview Guide}
\label{Guide}

{\scriptsize

\begin{longtable}{|p{1cm}|p{0.8cm}|p{1.5cm}|p{6.3cm}|p{2cm}|}
\caption{Interview guide on criteria for fairness certification of NLP approaches}

\label{tab:guide}
\endfirsthead
\hline
\textbf{Min} & \textbf{Min acc.} & \textbf{Goal} & \textbf{Contents and exemplary questions}  & \textbf{Expected} \par \textbf{results}\\
\hline
7 & 7 & Buffer, welcome  & Recent projects, travel,...   &  relaxed start  \\
\hline
3 & 10 & Sort out organizational matters & Obtain consent for recording, outline objective of work, outline process of interview, outline anonymization and tools used  & GDPR, research goal and procedure clear \\
\hline
5 & 15 & 
 Understand background of interviewee & In which step of the AI lifecycle are you primarily in contact with NLP systems?

What points of contact have you had with fairness so far?

What incidents or issues have there been with it? What contact points might there have been about certification?  & Contact points to topic, professional practice \par  of the interviewee 
\\

\hline
3 & 18 & Getting started with the topic  & What does fairness of AI or NLP mean to you? How would you define fairness of NLP?

What are your initial thoughts on the topic of fairness certification of AI? And NLP specifically? & Terminology clarified, entry \par point \\
\hline
10 & 28 & Criteria for \par  NLP system development & Which requirements should be made to processes in development to ensure fairness? Why? (Business understanding, data understanding, preprocessing, annotation, modeling, explainability, evaluation)

What are the organizational requirements that should be considered for fairness certification of NLP? Why? (Team, structure)

What must suppliers of software components adhere to in order to maintain fairness of the end product?

Do you see any other issues in the NLP systems development phase that have not yet been raised? What are the biggest challenges for NLP providers? Where are the opportunities? & Answers to \par  questions for development criteria; contextual \par understanding \\

\hline
10 & 38 & Criteria for \par NLP system\par  operation & 
    
Are there any aspects regarding deployment that should be considered?

How does an AI system need to be monitored after it goes live to detect fairness issues that arise? Who needs to be involved and how is the monitoring organized?

How often and how extensively do you think fairness problems should be checked? How might such an audit be conducted? What are the key issues that need special attention?

What processes should be initiated in case of fairness violations? How can improvements be made?

To what extent is the interaction of users with the system to be controlled? Why?

What interactions of the user with the system have to be considered?

What are opportunities and challenges for NLP providers? & Understanding how to maintain fairness in an operational system and why this is so, and how this can be captured in criteria 
\\
\hline

5 & 43 & Clear out \par  open topics & Are there any other criteria apart from the ones discussed so far that you would use for fairness certification in NLP?

What else is important to you to say about fairness certification in NLP? & Additional dimensions \\
\hline
2 & 45 & Pointing out next \par steps & Describe follow-up.

Do you know anyone else who could help me with an interview? & Further points \par  of contact \\

\hline

\end{longtable}}

%% file: AppendixDetailed.tex
\section{Detailed Interview Findings}
\label{sec:appendixeval}


\small
\subsection{Process Criteria}
Process Criteria target everything generally relevant to establishing a relevant fairness certification process. Interviewees describe concepts that can be clustered in the categories of fairness understanding, design of assessment, and certification market factors.
\subsubsection {Fairness Understanding}
Having a good understanding of the fairness properties for the given application can be seen as an essential foundation for building an effective certification process (\intvwr{12}, \intvwr{14}). Findings follow the structure of codes identified in open coding, as introduced in the previous chapter. It makes sense to keep perspective on the findings by first introducing subjective fairness understandings of interviewees. Some tended to focus on an equal opportunity aspect, others on equal performance or other conceptualization. The different perspectives of interviewees on fairness also may manifest in their suggestions for certification measures.
Interviewees one, four, six, and seven share the fairness perspective of Fairness through Awareness (\intvwr{1}, \intvwr{4}, \intvwr{6}, \intvwr{7}) which aims at providing similar outcomes to similar individuals \cite{dwork2012fairness}. This similarity establishes fair treatment on a meritocratic baseline (\intvwr{14}). Hence, factors that are objectively relevant to a decision, like skills, should be the baseline for a decision. Conditional statistical parity \cite{corbett2017algorithmic} could also be utilized as a suitable fairness conception from the literature. A form of enablement that may account for socioeconomic backgrounds, like migration (\intvwr{14}), could also be considered (\intvwr{14}). That would allow for the reduction of structural inequalities like gender discrimination in our society (\intvwr{6}). The target could be changing these structural inequalities by actively intervening (\intvwr{1}, \intvwr{4}, \intvwr{7}). This concept could be captured by statistical parity, where the demographics of groups that receive a positive (or negative) classification meet the demographics of the overall population \cite{dwork2012fairness}. This conception, however, comes with the limitation that a group's representation in the overall population needs to be large enough to be considered.
Treatment based on a meritocratic baseline is closely related to the notion of a non-discriminatory system. It avoids an unjustifiable negative impact on affected individuals (\intvwr{3}, \intvwr{5}) and is how interviewees three, four, five, ten, and 13 view fairness (\intvwr{3}, \intvwr{4}, \intvwr{5}, \intvwr{10}, \intvwr{13}). It closely resembles the literature's counterfactual fairness definition, which emphasizes decision invariance to a protected attribute \cite{kusner2017counterfactual}. It may depend on the use case in its context when utilizing a protected attribute to justify negative impact may be acceptable (\intvwr{3}, \intvwr{5}).
As mentioned by interviewees two, four, seven, eight, nine, and eleven, inclusiveness ties closely into this notion (\intvwr{2}, \intvwr{4}, \intvwr{7}, \intvwr{8}, \intvwr{9}, \intvwr{11}). It means a system should not be restricted in accessibility and usability, and there should not be a barrier to usage by a group (\intvwr{2}, \intvwr{11}). Ideally, the system also represents a diverse spectrum and avoids echo chambers (\intvwr{4}, \intvwr{7}, \intvwr{8}). There can also be trade-offs between fairness dimensions. For instance, striving for inclusiveness can impair fairness regarding counterfactual fairness in a recruitment context (\intvwr{1}).
Interviewees two, four, nine, and eleven emphasize equal performance as their subjective fairness understanding (\intvwr{2}, \intvwr{4}, \intvwr{9}, \intvwr{11}). It focuses on delivering the same performance to each user independent of group membership. The system can cope, for instance, with the variance of expression without sacrificing performance for certain groups (\intvwr{9}). It may even be interpreted as an equal user experience (\intvwr{9}). Literature also employs the mathematical fairness definition of counterfactual fairness to capture this conception of fairness \cite{kusner2017counterfactual}. For some use cases like voice assistants, this fairness conception can be even business-critical due to user diversity (\intvwr{2}, \intvwr{9}).
This subjective nature makes it more difficult to clearly define fairness in a certification process (\intvwr{1}, \intvwr{3}, \intvwr{5}, \intvwr{12}). Defining fairness before an assessment is an important factor for operationalizing a fairness certification (\intvwr{5}, \intvwr{8}, \intvwr{9}, \intvwr{12}, \intvwr{14}). Another challenge in defining fairness for an assessment is rooted in the strong dependence on context imposed by the use case (\intvwr{1}, \intvwr{3}, \intvwr{5}, \intvwr{6}, \intvwr{7}, \intvwr{8}, \intvwr{12}, \intvwr{13}, \intvwr{14}). This use case dependence is particularly highlighted by interviewees three, twelve, and 14. The use case may, for instance, determine whether procedural or distributional fairness should be the focus (\intvwr{14}). Required criteria for designing content for an assessment may vastly differ between fairness conceptions (\intvwr{6}, \intvwr{14}). Some interviewees see the large variance in use cases as a reason for calibrating an assessment to the specific use case by adapting its contents to the fairness situation the application imposes (\intvwr{6}, \intvwr{7}, \intvwr{11}, \intvwr{14}). It may make sense to employ different fairness definitions depending on the affected stakeholders and context for the certification process, as discussed in \enquote{Project Planning Criteria} for the audited company, and adapt the certification process accordingly (\intvwr{5}, \intvwr{8}). Moreover, a dynamic environment implies changing fairness perceptions over time, so fairness definitions for stakeholders may need to be reassessed (\intvwr{13}, \intvwr{14}). Ethics and moral considerations are closely related to fairness (\intvwr{8}, \intvwr{9}, \intvwr{10}, \intvwr{14}). A fairness definition can be grounded in the underlying definition of ethics (\intvwr{8}). Interviewee eight quotes Joscha Bach to define ethics as a \enquote{[r]ational process negotiation with yourself and others to decide how to resolve a conflict of interest under the condition of shared value} (\intvwr{8}). Hence, grounding a fairness definition in ethics could cope with the varying fairness conceptions (\intvwr{8}).
An assessment should ideally not result in a binary outcome in the form of fair vs. not fair, but it should be more nuanced, describing to what degree which fairness concepts are met (\intvwr{12}). Otherwise, it is difficult to determine the point when something starts to become unfair (\intvwr{8}). It is crucial to view fairness as a societal issue as our fairness perception is culturally rooted in the same way some biases are (\intvwr{6}, \intvwr{12}). Different values in another cultural context may cause a regional dependence of fairness (\intvwr{9}, \intvwr{13}, \intvwr{14}). The fairness perception in China may not align with the typical European fairness understanding, for instance (\intvwr{13}). This societal bias poses a challenge for assessing and operating models on a global scale (\intvwr{13}).
In practice, the topic of fairness in NLP systems is barely considered by most development teams (\intvwr{1}). That partially leads to a lack of understanding and awareness of the problems that must be addressed in a certification process (\intvwr{11}, \intvwr{14}). Besides a lack of practical understanding of creating a good fairness certification process, some fundamental insights from research to ground such a process are still missing (\intvwr{14}). Fairness should not be viewed as an isolated concept as it is entangled with other variables of Responsible Artificial Intelligence (RAI) (\intvwr{1}, \intvwr{7}, \intvwr{9}). Transparency, for instance, has already been addressed as a substantial part of a fairness certification. Data security, privacy, and reliability are also named as influencing factors on fairness (\intvwr{1}, \intvwr{7}, \intvwr{9}). Improving such should not affect fairness negatively but rather benefit it, making an integrated assessment of fairness with other RAI variables interesting (\intvwr{9}).
\subsubsection {Design of Assessment}
\begin{figure*}[htbp]
   \centering
   \includegraphics[width=1\textwidth]{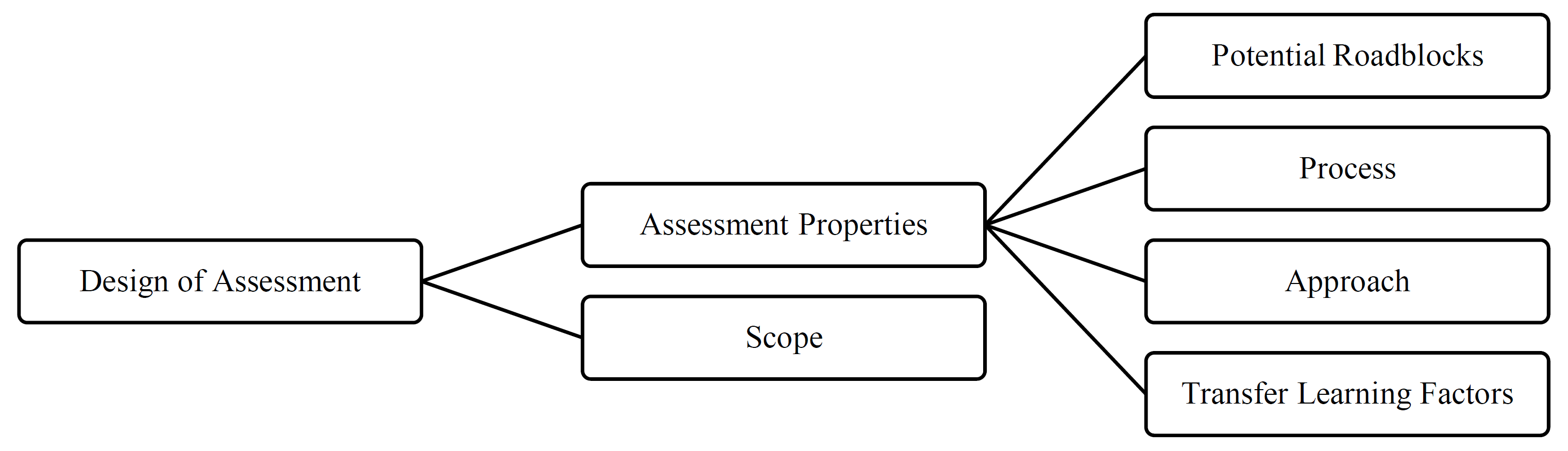}   
   \caption{Criteria relevant to \enquote{Design of Assessment} hierarchically mapped}
   \label{fig:3}
\end{figure*}

Interviewees also give insight into what they consider a suitable design of an assessment for certifying fairness. That includes general properties of the certification process that the certifying institution should consider and information about the scope of an assessment. The hierarchical open coding scheme for \enquote{Design of Assessment} is displayed in Figure \ref{fig:3}.

\textit{Assessment Properties}:
Assessment properties describe how an assessment should be approached and what may be considered when creating such an assessment procedure. However, potential roadblocks need to be considered for an assessment to gain traction. First, the feasibility of such a holistic certification process that does not induce unfairness among industry players is questioned to a certain extent (\intvwr{3}, \intvwr{5}, \intvwr{6}, \intvwr{11}). Apart from the resource intensity of undergoing such a certification process which makes it inaccessible for small players (\intvwr{1}, \intvwr{13}), there is the potential issue of only limited access which can be granted to the auditor by the company (\intvwr{12}, \intvwr{13}) and the issue of complexity of the topic as well as the systems (\intvwr{3}, \intvwr{5}) which may make it unattainable in the near future (\intvwr{14}). A certification process cannot be seen as a guarantee that no bias issues will occur (\intvwr{5}). A challenge to master for the auditor is to provide an assessment that is as objective as possible, which is difficult with humans involved (\intvwr{13}).
The approach may define how effective a certification process turns out to be. Currently, approaches to auditing the fairness of a system tend to be too selective and hence do not cover the entire range of problems that need to be addressed (\intvwr{14}). Having an end-to-end approach covering the system holistically may be essential (\intvwr{1}, \intvwr{6}, \intvwr{8}, \intvwr{9}, \intvwr{11}, \intvwr{12}, \intvwr{13}, \intvwr{14}). Also, hard-coded parts of the system should be investigated in an assessment as they sometimes can be decisive for the system’s outputs (\intvwr{6}, \intvwr{9}). It may make sense to focus on an assessment regarding the most pressing fairness issues so that a system can operate without causing anyone considerable harm (\intvwr{1}, \intvwr{2}, \intvwr{3}, \intvwr{10}). An approach that is partially practiced in the industry employs a human comparison to define the threshold for unfairness (\intvwr{4}). It may be criticized as it does not reflect the target of systems’ being as fair as possible (\intvwr{2}, \intvwr{4}). Another approach focuses more on certifying individuals involved in the systems development and operations than on processes established in an organization or data and reports generated by it (\intvwr{5}, \intvwr{6}, \intvwr{8}). Interviewee eight describes the concept of quick checks, which is discussed further in the chapter \enquote{Criteria for Governance}, as such an employee-focused certification process may be more effective (\intvwr{8}) and may have a longer-lasting lifespan in ensuring fairness (\intvwr{6}). Another perspective envisions a modular certification process where different endeavors, like anonymizing data, are all certified individually (\intvwr{5}).
The process factors define specific nuances a certification process should consider and what is required for implementation. The assessment process should first differentiate between assessment for Business-to-Business (B2B) and assessment for Business-to-Consumer (B2C) use cases (\intvwr{8}). In B2B use cases, different aspects like competitive fairness are relevant, which are irrelevant for B2C and the other way around (\intvwr{4}, \intvwr{7}). An assessment should be built on standards, KPIs, and best practices (\intvwr{1}, \intvwr{5}, \intvwr{8}, \intvwr{10}, \intvwr{11}, \intvwr{13}, \intvwr{14}). They help companies align their processes and offer clarity. Existing frameworks should be utilized to build a process on (\intvwr{1}, \intvwr{8}, \intvwr{12}, \intvwr{13}). Currently, only internal certification processes are established in some companies, which generate Lessons Learned (\intvwr{14}). The latter eventually translates over time into the development of an independent third-party assessment facilitated by an exchange of discovered working mechanisms between practitioners and in exchange with policymakers (\intvwr{14}). Currently, the development of processes for auditing fairness is driven by assessing the practical applicability of research (\intvwr{14}). When an independent third-party certification is first introduced, it will also manifest and improve with learnings out of experimentation and practical application (\intvwr{1}, \intvwr{13}, \intvwr{14}). When standards are set and a consistent process is established, the reasoning for reaching the result of an assessment is to be addressed, as it should be grounded in a factual foundation (\intvwr{13}).
Transfer learning needs to be considered separately here as the provider of the unspecified pretrained baseline model and the provider of the custom model, which is then trained for the specific task, are involved (\intvwr{2}, \intvwr{9}, \intvwr{11}). Interviewee nine argues that the certifications of both models should be kept distinct so that the custom model is certified independently of the ideally already certified baseline model (\intvwr{9}). That may entail that only already certified baseline models are safe and viable options when pursuing certification for one’s custom model (\intvwr{11}). Reviewing the baseline model by oneself as a provider of a custom model is barely feasible (\intvwr{11}). Some interviewees argue that with training the custom model (\intvwr{2}, \intvwr{9}) or due to the use case dependence of fairness (\intvwr{1}), there is a responsibility shift regarding fairness from the provider of the baseline model to the provider of the custom model. That is because the data for training on the baseline model defines the final behavior of the model regarding fairness (\intvwr{1}, \intvwr{2}, \intvwr{9}).

\textit{Scope}: 
Interviewee nine poses the critical question of where the scope of a certification process should end (\intvwr{9}). Nearly all interviewees discuss the scope of a certification process to a certain extent (\intvwr{1}, \intvwr{2}, \intvwr{3}, \intvwr{4}, \intvwr{5}, \intvwr{6}, \intvwr{7}, \intvwr{8}, \intvwr{9}, \intvwr{11}, \intvwr{12}, \intvwr{13}, \intvwr{14}). It may make sense to start basic first when introducing a certification process for fairness and grow it more nuanced and complex over time (\intvwr{9}, \intvwr{14}). The generalization of the certification process is a major concern regarding its scope (\intvwr{3}, \intvwr{5}, \intvwr{7}, \intvwr{8}, \intvwr{9}, \intvwr{14}). A certification should be as general as possible without sacrificing the ability to properly judge a use case (\intvwr{3}, \intvwr{14}). There is a need to specify criteria depending on which use case is present (\intvwr{3}, \intvwr{5}, \intvwr{8}, \intvwr{11}, \intvwr{12}, \intvwr{13}, \intvwr{14}) or the geographical location the model is used (\intvwr{9}). As an example, a hiring context and a context in the medical domain may be compared (\intvwr{14}). In the hiring context, it makes sense to introduce criteria assessing the similarity of embeddings between genders in the semantic vector space. In the medical domain, the differences between genders may be valuable information to improve treatment specifically for each gender. Criteria should not favor semantic similarity there, but rather focus on the result, which is the success of the treatment for each gender. Clustering into similar use cases may be problematic as human perception of similarity may not align with what models represent (\intvwr{3}), and there needs to be more clarity for assigning clusters (\intvwr{14}). Another challenge when specifying criteria is determining what is to be considered relevant (\intvwr{3}).
The scope that should be certified may also vary in its thoroughness. That may depend on either the risk invoked by the system regarding unfairness (\intvwr{1}, \intvwr{2}, \intvwr{6}, \intvwr{9}, \intvwr{14}) or on the number of users affected by it (\intvwr{3}). The risk of a system is typically rooted in its application and not just in the underlying technical system and hence may shift with the application context changing (\intvwr{8}). For a risk-based differentiation between use cases, it may make sense to hierarchically increase the thoroughness and frequency of an assessment with the societal impact a system has (\intvwr{6}, \intvwr{14}). For low-risk applications, interviewee 14 suggests that no external assessment may be required, and an internal assessment should be pursued by the company and documented (\intvwr{14}). That way, internal product, and process expertise can be leveraged in an internal audit to make it more streamlined (\intvwr{14}). However, an external assessment should be regularly required for high-risk applications (\intvwr{14}). It may also make sense to require more transparency with increasing risk (\intvwr{14}). To assess the risk of an application, systemic issues like gender discrimination should be considered. This may be done by looking at the potential severity of impact in isolation and the cumulated effect of combined negative impacts in the context of our societal system (\intvwr{14}). Ultimately, it is a societal consideration how much emphasis is put on certification (\intvwr{14}).
To push forward fairness in the industry, some interviewees suggest making a certification at least in some cases obligatory as they fear lacking utilization of certification without an obligation (\intvwr{2}, \intvwr{10}, \intvwr{11}). Particularly, large language models are mentioned to require a certification process (\intvwr{4}). Others argue that certification should much rather provide guidelines based on which the application is certified without introducing an obligation (\intvwr{3}, \intvwr{5}). Some interviewees suggest that it may make sense to certify a more extensive scope than NLP fairness (\intvwr{8}, \intvwr{9}, \intvwr{14}). A certification for ML or even algorithmic fairness, in general, is suggested (\intvwr{8}, \intvwr{9}), and an extension of the certification process to assess other RAI principles is proposed (\intvwr{9}).
\subsubsection {Certification Market Factors}
Market factors describe how a certification process would influence its major stakeholders to act economically and how the market environment is shaped. Currently, uncertainty is in the market as there are no established standards or best practices and regulatory uncertainty (\intvwr{1}, \intvwr{14}). There is a reciprocal influence between a certification process and potential future regulation regarding systems’ fairness. On one side, the prospect of coming regulation in the field incentivizes companies to prepare for future compliance by already undergoing certification (\intvwr{1}, \intvwr{3}, \intvwr{14}). On the other hand, certification may guide regulation by introducing standards and giving direction to industry players (\intvwr{4}, \intvwr{6}). An important factor for adopting a certification process lies in its alignment with the company’s business goals and viability (\intvwr{3}, \intvwr{11}). The company first needs to accept the fairness criteria, which are assessed by the certification process (\intvwr{9}). It also needs to see the potential benefits of a certification process. A fairness assessment process may make the company less vulnerable to potential lawsuits (\intvwr{1}, \intvwr{7}), signal trustworthiness and reliability to the user (\intvwr{8}, \intvwr{9}), and even drive innovations (\intvwr{3}, \intvwr{9}). Fairness issues become increasingly critical (\intvwr{9}), and certification may be an important selling point for a system (\intvwr{1}, \intvwr{10}, \intvwr{11}, \intvwr{14}). The more business-critical fairness becomes, the higher the certification adoption rate (\intvwr{13}). A whole market may develop around such a certification of fairness, as most companies will require external consulting to ensure compliance with established standards and certification requirements (\intvwr{1}).

\subsection{Governance Criteria} 
Governance Criteria focus on governance aspects that impact a system's fairness and should be assessed in a fairness certification process. A certification process may be beneficial as it can augment the company's internal assessment procedures and provide feedback that is, in the long term, beneficial for the company (\intvwr{7}). In the context of fairness, governance should establish a moral standard and a reciprocal fairness awareness, as users and providers should be able to judge fairness appropriately (\intvwr{3}, \intvwr{4}, \intvwr{8}). This may be established by introducing a certification process, as it is currently barely considered in the industry (\intvwr{10}). However, there may be one constraint regarding governance if a company is developing systems for customers who will use them for their products. The solution provider then needs to satisfy the customer's wishes and hence is restricted regarding governance (\intvwr{6}, \intvwr{10}). The interviews brought to attention two major concepts that need to be considered for governance: \enquote{Model Reporting \& Transparency} and \enquote{Organizational Criteria}.
\subsubsection {Model Reporting \& Transparency}

\begin{figure*}[htbp]
   \centering
   \includegraphics[width=1\textwidth]{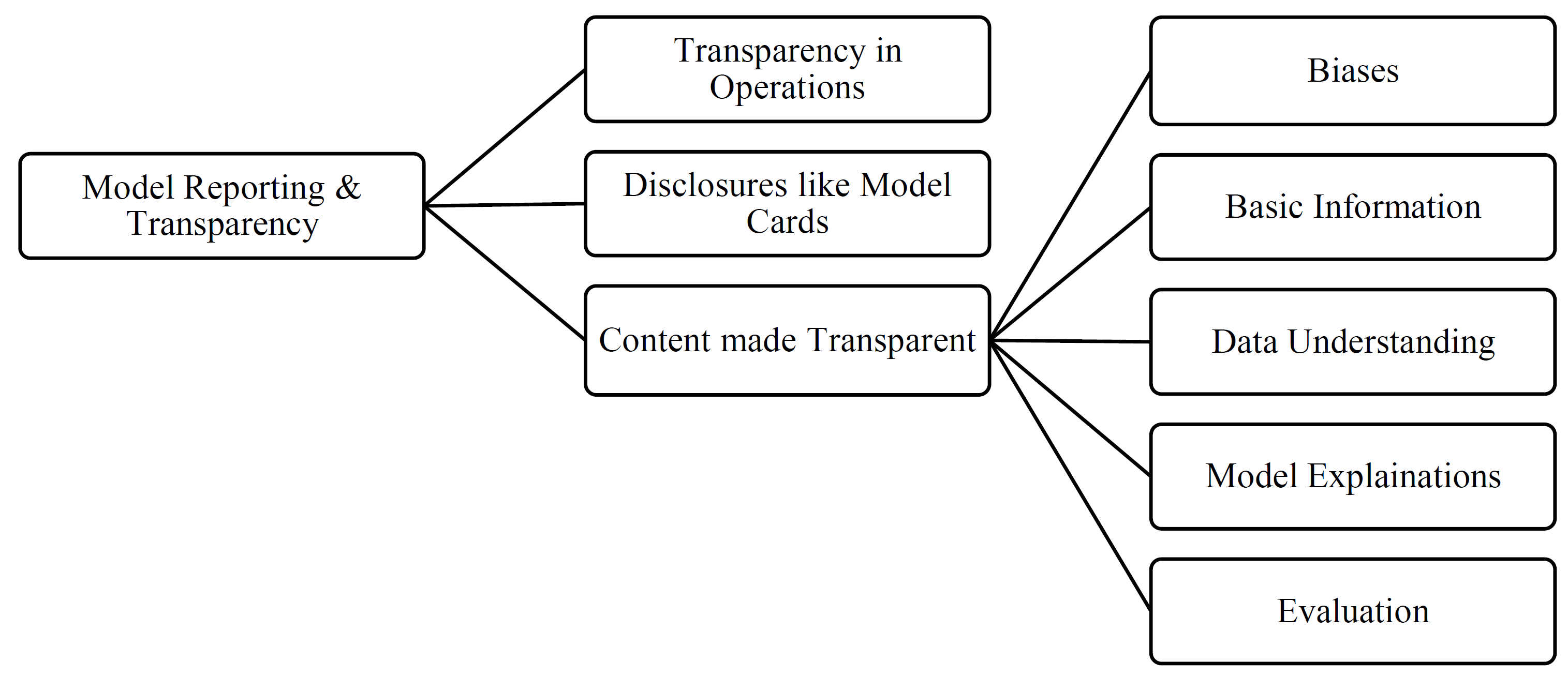}   
   \caption{Criteria relevant to \enquote{Model Reporting \& Transparency} hierarchically mapped}
   \label{fig:4}
\end{figure*}

Providing model reporting and transparency focuses on reducing information asymmetries between the system’s provider and stakeholders by providing information about the application that may be relevant to stakeholders, particularly for identifying biases (\intvwr{1}, \intvwr{3}, \intvwr{6}, \intvwr{11}, \intvwr{13}) or avoiding the occurrence of fairness issues. It can be seen as an accountability mechanism for the system’s provider (\intvwr{1}, \intvwr{7}, \intvwr{13}). Model reporting may also provide essential information for an audit regarding fairness (\intvwr{4}, \intvwr{13}). Criteria for model reporting are hierarchically mapped in Figure \ref{fig:4}.
Transparency may be essential in maintaining a good reputation as a company (\intvwr{7}). It may be limited by confidentiality requirements within the company (\intvwr{13}). The relevance of good model reporting and transparency may be particularly high for API services (\intvwr{2}). That is because the customer using the service or building on it faces information asymmetry and only accesses the system via an interface (\intvwr{2}). The form in which transparency is provided may also vary.
The ideal form of transparency in operations would be given with open-sourcing key components of the system (\intvwr{8}). That enables stakeholders to improve components of the system as they can not just understand the system but also participate (\intvwr{8}), which may be beneficial regarding fairness as affected groups actively can contribute. That way, their perspective might be better represented in the system’s continuous improvement. A lower degree of stakeholder access may be chosen with only information about the system and its inferences provided as another form of transparency in operations. Then there need to be considerations regarding the design of the interface providing such information as it should not negatively impact the system's usability (\intvwr{8}). Users may not all be interested in transparency about the system (\intvwr{8}). Interviewee eight suggests making such information immediately available with the click of a button (\intvwr{8}).
An often-mentioned implementation of model reporting can be found in Model Cards or closely related conceptions (\intvwr{2}, \intvwr{3}, \intvwr{4}, \intvwr{8}, \intvwr{12}, \intvwr{13}, \intvwr{14}). The paper introducing model cards \cite{mitchell2019model} is mentioned and referred to by interviewees (\intvwr{12}, \intvwr{13}, \intvwr{14}). A model card reports on model details, intended use, factors that may influence the model regarding fairness, metrics evaluation data, training data, quantitative analyses, ethical considerations, and caveats and recommendations \cite{mitchell2019model}. The version of model cards provided by Hugging Face \cite{wolf2019huggingfaces} is also mentioned in the context of providing transparency (\intvwr{2}, \intvwr{3}). Interviewee eight suggests an extended and adapted concept (\intvwr{8}) which inspires an adapted version extended with some contents mentioned in this chapter and can be found in Table 5 in the Appendix. Introducing a similar concept to model cards leads to better-documented models and improved transparency, which are beneficial, as previously mentioned, not just for fairness (\intvwr{3}, \intvwr{13}). Model cards can help by giving direction to model users or even providers by raising awareness on fairness issues and potential risks of an application and starting a reflection process (\intvwr{8}).
The content of disclosures in operations or model reporting can be clustered in biases, basic information, data understanding, model characteristics, and evaluation. Basic information should include information about the intended fields of application (\intvwr{4}, \intvwr{13}), the underlying fairness definition, and the frameworks the system was built on (\intvwr{4}). The latter may be important information for an audit, making the process reproducible (\intvwr{4}). Moreover, information about the model's generalizability in the domain should be provided, giving an indication of model quality (\intvwr{4}). Furthermore, it should be communicated by the company based on what fairness definition it built the system and what the latter excludes (\intvwr{12}).
Data understanding covers all relevant information about the data the system utilizes and how it is processed. This is essential as NLP systems tend to be trained on vast corpora of sometimes even confidential data, which makes it hard to judge for users what fairness issues may potentially arise (\intvwr{13}). Data understanding involves providing the context of data collection, as there might be incentives for unrepresentative data collection (\intvwr{12}). Moreover, the data source (\intvwr{12}), the volume of data (\intvwr{4}) as well as distributions (\intvwr{12}) should be made transparent. It should become clear what has been used as training data and what as evaluation data (\intvwr{4}). Data splitting and sampling should be described, and their representativeness stated (\intvwr{4}). All data preprocessing steps should be described (\intvwr{12}) and a sample of the data may be provided (\intvwr{13}).
Transparency may also be important regarding model explanations (\intvwr{1}, \intvwr{2}, \intvwr{3}, \intvwr{4}, \intvwr{6}, \intvwr{7}, \intvwr{8}, \intvwr{11}, \intvwr{12}). It should first be reported which models are utilized for the system (\intvwr{4}, \intvwr{12}). It may also make sense to go deeper and document how the model trains itself on all layers (\intvwr{8}). This ties into how a model learns certain things, which is addressed by explainability (\intvwr{11}). Explainability should allow the user to understand the basis of the system’s decision and contextualize it (\intvwr{1}, \intvwr{2}, \intvwr{3}, \intvwr{7}, \intvwr{8}). It may be essential to make explainability information transparent (\intvwr{2}, \intvwr{3}, \intvwr{4}, \intvwr{6}, \intvwr{8}, \intvwr{12}). The users’ domain expertise in the specific field may help uncover particular fairness issues (\intvwr{6}). Explainability enables a plausibility check by a human to see whether unwanted decision drivers may be involved that could indicate fairness issues (\intvwr{2}, \intvwr{3}, \intvwr{4}). In some cases, explainability may also be a business requirement and needed anyway (\intvwr{4}). However, there may be a tradeoff in the quality of explainability feasible for a model and its performance. That is because large and complex models that make operationalizing explainability difficult tend to perform better (\intvwr{2}, \intvwr{3}, \intvwr{4}). Smaller, interpretable models may be a solution for certain tasks (\intvwr{4}, \intvwr{6}), as discussed in the chapter \enquote{Project Planning Criteria}. Multiple explainability tools are utilized for black-box models for which a minimum standard that needs to be met may be specified in a certification (\intvwr{2}). But it should consider the different suitability of certain tools depending on the use case (\intvwr{2}). Mentioned tools that may be utilized include attention scores over all layers (\intvwr{2}), integrated gradient score and token influence prediction (\intvwr{2}, \intvwr{4}), and visualization (\intvwr{3}, \intvwr{8}). The latter may be hard to quantify (\intvwr{3}). Non-quantifiable measures involve a process of human sensemaking, which induces a bias (\intvwr{3}). Because of that, quantifiable explainability measures should be preferred (\intvwr{3}). The last consideration for reporting explainability is the comprehensiveness of the communicated information to non-expert users (\intvwr{8}, \intvwr{12}). Just reporting very technical details on the model and its inference is probably useless for most users, and the addition of comprehensive explanations is required (\intvwr{8}, \intvwr{12}).
The steps taken for evaluation should also be reported. It should be reported what has been tested regarding the robustness of the model against biases (\intvwr{2}, \intvwr{4}). There may be some form of scoring on standardized test sets targeting various biases and allowing for a benchmarking of the model, which can be reported (\intvwr{2}, \intvwr{4}, \intvwr{13}). Evaluation metrics should be publicly documented and reported (\intvwr{4}, \intvwr{8}).
The reporting of biases covers data bias (\intvwr{4}), model bias (\intvwr{3}), and countermeasures against biases that are implemented (\intvwr{3}, \intvwr{13}). If biases can’t be mitigated, there should be at least a hint to users of the system priming them for the potential occurrence of such (\intvwr{4}, \intvwr{7}). Interviewee two also suggests testing on tasks not intended for the system but may give hints regarding bias issues that may also occur for the original task and hence should be reported (\intvwr{2}).
\subsubsection {Organizational Criteria}
\begin{figure*}[htbp]
   \centering
   \includegraphics[width=1\textwidth]{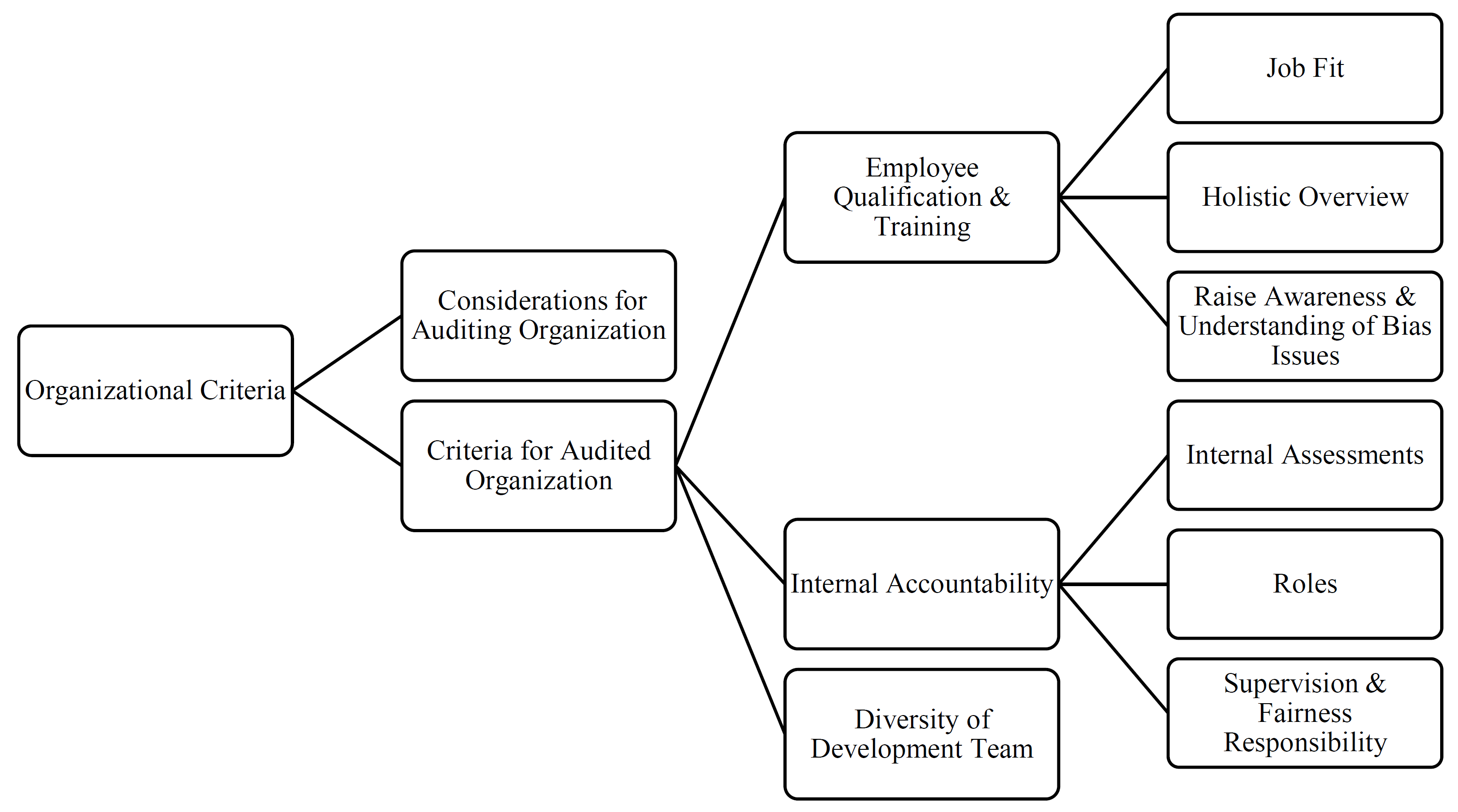}   
   \caption{\enquote{Organizational Criteria} hierarchically mapped}
   \label{fig:5}
\end{figure*}

Organizational criteria revolve around general processes established in an organization, accountability-related aspects, and team- and qualification-related aspects. The findings of this work include organizational criteria for the audited and considerations for the auditing organization. The hierarchical structure of all mentioned concepts is visualized in Figure \ref{fig:5}.

\textit{Considerations for the auditing organization} mainly focus on who is performing the assessment. When first establishing a certification process, it may make sense to collaborate with independent auditing firms with expertise (\intvwr{12}). It may also be important to consider the background of the certifying third party as the region it is located in (\intvwr{13}), the political system it is integrated into (\intvwr{13}), its initiator (\intvwr{13}), or its integration in the economic system (\intvwr{2}, \intvwr{10}) influence it. Interviewee 13 suggests an international union like the OECD may be a good platform to build such a certification process (\intvwr{13}). Experts should be responsible for the certification assessment process and continuously consolidate and update best practices (\intvwr{2}, \intvwr{6}, \intvwr{9}, \intvwr{12}). Regarding the background of those experts, interviewees suggest involving linguists (\intvwr{9}), domain experts (\intvwr{6}), data scientists and data engineers (\intvwr{6}), and ethics or sociology experts (\intvwr{6}, \intvwr{9}).

\textit{Criteria for the Audited Organization}:
Organizational criteria for the audited organization include employee qualification and training, internal accountability, and diversity of the development team. For employee qualification and training, the job fit of the employee is an important concept (\intvwr{3}, \intvwr{6}). In a step like annotation, there can be the issue that due to cost reasons, typically, employees assigned to annotate are not on long-term contracts and hence may also lack the motivation to make a conscious effort to follow annotation guidelines (\intvwr{6}). The latter, however, is crucial for fairness as it provides the foundation to train the model (\intvwr{6}). Who is assigned what task may indicate how successfully fairness may be implemented (\intvwr{6}). Hence, it would make sense to assess who is assigned what job and whether qualifications are sufficient to avoid imposing fairness risks (\intvwr{6}).
The awareness for fairness needs to be raised, and an understanding of bias issues needs to be present in the organization as there tends to be lacking expertise regarding fairness in development teams (\intvwr{1}, \intvwr{3}, \intvwr{4}, \intvwr{6}, \intvwr{7}, \intvwr{11}). That needs to be done for the development team of an application by sensitizing to potential biases and training (\intvwr{1}, \intvwr{3}, \intvwr{6}, \intvwr{7}). Again, annotators are highlighted as particularly relevant for such training (\intvwr{3}, \intvwr{6}). Moreover, customer-centric teams should be trained in detecting and identifying biases (\intvwr{11}). That may be achieved by establishing practices that make customer-centric teams build empathy with the affected individuals and change their perspective to understand the affected groups' perceptions (\intvwr{11}). Furthermore, such practices may be essential when first introducing a certification process as they enhance the overall problem understanding, which in turn helps the certification process to evolve (\intvwr{11}). The installment and contents of such training might be subject to an audit.
Moreover, members of the development team should all have a rough holistic overview (\intvwr{6}, \intvwr{8}). A possible measure supportive of this would be offering workshops that foster a holistic overview (\intvwr{6}). It may be checked that every development team member is aware of their fairness impact on the product (\intvwr{8}). Contextualizing such quick checks for different roles that each have a different perspective on the product may be a valuable approach to fostering fairness (\intvwr{8}). Tech-focused team members may benefit from a quick check on data understanding and mitigation layers (\intvwr{6}, \intvwr{8}). Annotators and non-tech-focused team members can be equipped with a system and process understanding (\intvwr{6}, \intvwr{8}). That would shift the focus of the certification audit further onto the employees as their workflow is certified, similar to a checklist (\intvwr{8}).
For internal accountability, roles need to be assigned clear responsibilities, internal assessment or audit procedures should be established, and supervision and responsibility for fairness must be addressed. Starting with roles, establishing a fairness officer or team dedicated to fairness makes sense (\intvwr{1}, \intvwr{2}, \intvwr{6}, \intvwr{7}, \intvwr{9}, \intvwr{10}, \intvwr{14}). That person or team should be responsible for the fairness definition in the specific use case (\intvwr{1}, \intvwr{9}), conducting an internal fairness assessment (\intvwr{1}, \intvwr{2}, \intvwr{6}) and providing advice on fairness to the development team and supporting it, for instance in its decision making (\intvwr{1}, \intvwr{7}, \intvwr{9}). The main issue that arises is that such an individual or team requires financial resources (\intvwr{2}, \intvwr{11}, \intvwr{14}). There may also be an external assessor for fairness who reports to the company (\intvwr{1}). The development team functions as a problem solver for fairness issues reported by one of the above assessors (\intvwr{1}, \intvwr{3}, \intvwr{11}) without being the responsible instance for detecting fairness issues (\intvwr{1}, \intvwr{6}, \intvwr{11}). It may also make sense for certain use cases to implement different levels of clearance, for instance, when confronted with sensitive contents in a feedback loop to ensure appropriate and fairness-compliant handling (\intvwr{7}).
For the specific project, interviewees view some form of supervision or fairness responsibility as an essential concept. Either the data curator (\intvwr{3}), a separately consulting organizational instance (\intvwr{7}), or the project or department manager (\intvwr{6}, \intvwr{11}) should be responsible for fairness in the specific project. The initiative for fairness, however, needs to come from the company’s particular commitments regarding fairness (\intvwr{11}). Some processes, like annotation, also need careful supervision and verification by someone accountable (\intvwr{4}, \intvwr{6}).
An internal audit or assessment can also be a vital tool for a company to ensure accountability regarding fairness (\intvwr{7}, \intvwr{14}). An assessment may target the impact an application has on its stakeholders as a foundation for fairness considerations (\intvwr{7}). It may be updated and tracked continuously, and with changing stakeholders, it needs to be reiterated (\intvwr{7}). Red teaming is suggested as another form of assessment (\intvwr{7}). It provides a holistic system testing for potential issues in a set time interval. The so-called \enquote{red team} is to find fairness vulnerabilities in the system. That could be seen as an exemplary holistic and focused company internal fairness assessment. Such assessments are typically motivated by ensuring to be regulatory compliant with future legislation besides ensuring fairness (\intvwr{14}). These assessments or internal audits should take place continuously (\intvwr{7}) and should not be conducted by the development team but rather by a separate organizational instance (\intvwr{14}). An aspect that may be assessed which may vary in its importance for fairness between use cases, is the amount of human oversight in operations (\intvwr{4}). One more dynamic should be considered regarding accountability which addresses binding claims made by a supplier about his product that shift responsibility back to the supplier (\intvwr{1}).
An assessment should also check the diversity of the development team (\intvwr{2}, \intvwr{11}). Due to different perceptions and realities of life, a diverse development team fosters problem-understanding and thus helps identify and solve fairness issues (\intvwr{2}). The major challenge lies in defining what a diverse team is (\intvwr{2}). It is debatable which dimensions like gender, race, or sexual orientation diversity are relevant to consider when building a diverse team (\intvwr{2}). It should ideally represent minorities (\intvwr{2}). Currently, development teams lack the diversity of their users in dimensions like age and gender (\intvwr{2}, \intvwr{11}). To be, to a certain degree, representative of the user may be a requirement to check (\intvwr{2}, \intvwr{11}). Diversity may also be supported via (sometimes lacking) interdisciplinary collaboration within the development team (\intvwr{6}, \intvwr{10}). Interviewees particularly highlight diversity in the development team for annotators as they directly impact and potentially bias data (\intvwr{4}, \intvwr{10}, \intvwr{13}). That, however, comes with the limitation that someone needs to be qualified to label, which may be more or less restrictive, dependent on the use case (\intvwr{10}).

\subsection{Project Planning Criteria} 
Project planning criteria are set up to assess the considerations regarding fairness in the project’s planning phase, where business understanding and process planning are done, and requirements for the project are set. Interviewees are divided on whether project planning criteria are useful to implement and what should be assessed with what kind of purpose in mind. In such early stages of the lifecycle, one interviewee argues it might be challenging to set criteria for certifying fairness (\intvwr{13}). Another interviewee argues that a consequence of criteria for business understanding may be neglecting some products’ specialized nature, leaving little room for steering decisions (\intvwr{6}).
\subsubsection {Define \& Assess the Planned Application}

\begin{figure*}[htbp]
   \centering
   \includegraphics[width=1\textwidth]{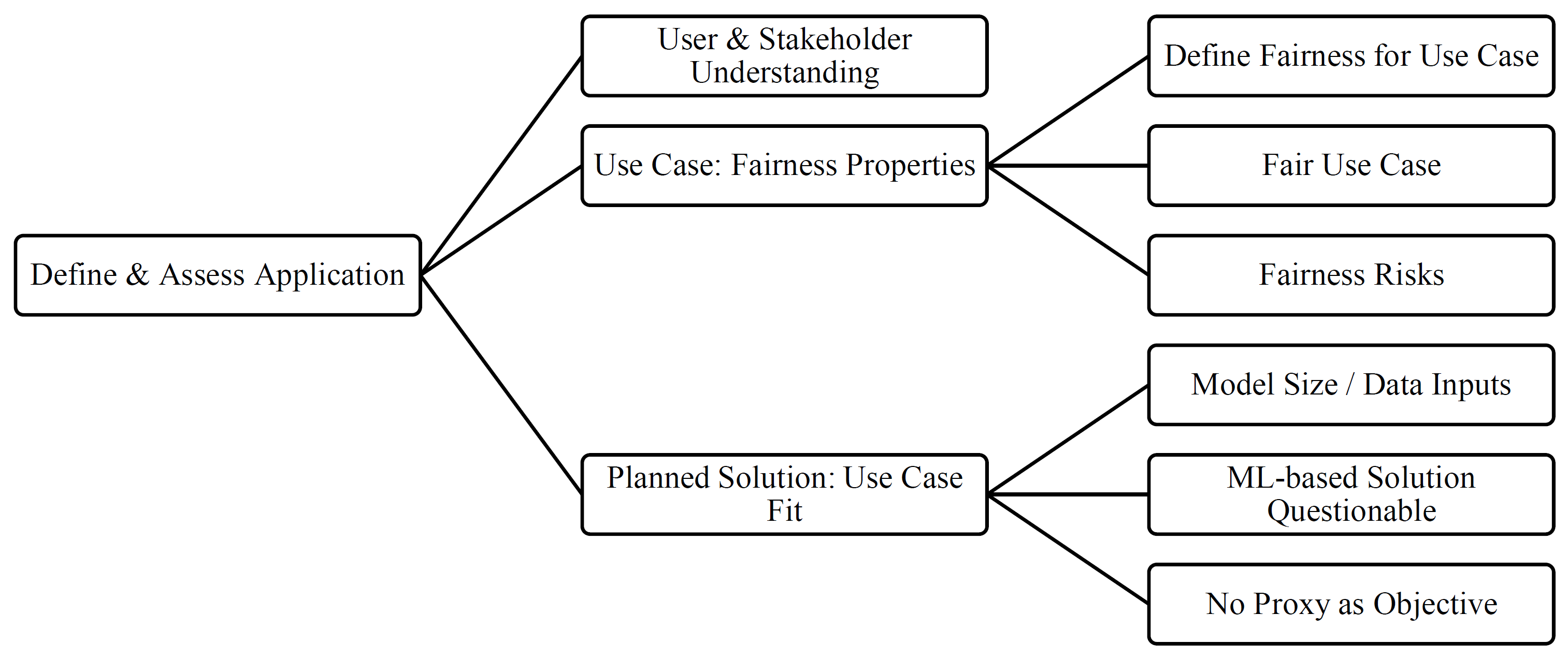}   
   \caption{Criteria relevant to \enquote{Define \& Assess Application} hierarchically mapped}
   \label{fig:6}
\end{figure*}

Interviewee 14 argues that in a certification process, the purpose of use and not the technology itself should be the focus of an assessment for certification (\intvwr{14}). Hence, the problem definition that the system builds on should be investigated (\intvwr{8}, \intvwr{12}, \intvwr{13}). That motivates an assessment that can already occur before the actual implementation starts, even though it may be associated with many manual processes (\intvwr{11}). 
According to many interviewees, solid definitions of the use case and the application planned are essential factors that should be checked in a fairness assessment (\intvwr{4}, \intvwr{8}, \intvwr{9}, \intvwr{12}, \intvwr{13}, \intvwr{14}). Moreover, users or stakeholders, in general, need to be identified (\intvwr{8}). In pursuit of being able to assess the impact on stakeholders and the planned application, three major concepts need to be considered: First, a good user and stakeholder understanding needs to be established. Second, the fairness properties of the use case need to be identified and assessed. Third, the planned solution must be checked to determine whether its fit for the use case favors fairness. The hierarchical mapping of all relevant concepts can be found in Figure \ref{fig:6}.

\textit{User \& Stakeholder Understanding}:
User and stakeholder understanding capture assessing processes to gain knowledge about general attributes and characteristics of stakeholders like demographics (\intvwr{6}), a target group definition (\intvwr{8}), the understanding of possibly multiple levels of distinct users involved (\intvwr{8}), and the context users find themselves in (\intvwr{11}). The assessment of the context of the users is highlighted by interviewee eleven as some stakeholders may be dependent on the system without having alternatives (\intvwr{11}), making a fairness issue more impactful. Interviewee eight highlights the importance of an assessment for direct and indirect users as well as other stakeholders separately, instead of a singular focus on the end user, to avoid vague results (\intvwr{8}). This conception is supported by the fact that some systems operate in multi-sided markets, and fairness impacts multiple stakeholders (\intvwr{7}).
The composition of the target group should be explained for an assessment to check (\intvwr{7}, \intvwr{8}). Some use cases may naturally exclude certain groups without being unfair (\intvwr{7}). Moreover, the audited company should have mechanisms to assess the likelihood of marginalized groups within the individuals affected by the application (\intvwr{8}). It should be assessed whether the planned system considers the particular interests of all its user groups (\intvwr{11}). However, it needs to be checked whether such processes may lead the company to develop misconceptions of their stakeholders as there is a risk of misinterpretations about their true nature (\intvwr{9}). 

\textit{Use Case: Fairness Properties}:
Assessing the fairness properties of the use case involves checking the definition of fairness for the use case and the implications of the latter on the defined stakeholders (\intvwr{12}, \intvwr{14}). A definition can be seen as essential as fairness aspects can differ drastically between use cases and conceptualizations of fairness (as discussed in the previous chapter). It sets the focus of what needs to be considered for the technical system (\intvwr{14}). As different stakeholders are involved, there may be multiple fairness definitions to consider (\intvwr{12}). What can also be assessed in an audit is whether the fairness approach that the audited company chooses is suitable for the use case (\intvwr{14}).
The underlying business case should also be assessed regarding fairness (\intvwr{9}, \intvwr{10}, \intvwr{14}). An unfair underlying business case invalidates all attempts to create a fair system (\intvwr{9}, \intvwr{10}). Interviewee 10 suggests utilizing a question catalog with yes/no questions on the use case. The business responsible for the project may fill it out and provide it for an audit to identify the fairness of a use case (\intvwr{10}).
Another property of a use case that needs to be considered is its vulnerability to fairness risks which may be introduced via sensitive criteria (\intvwr{7}). Name (\intvwr{5}) and region (\intvwr{2}, \intvwr{5}) are just two examples of such attributes. Credit scoring is mentioned as an exemplary application strongly affected by fairness risks (\intvwr{13}). An assessment should not just check whether the fairness risks of a use case are being identified but also if there are mitigation strategies for those risks in place (\intvwr{7}). The last property to consider is how much fairness is a viable business factor for the use case (\intvwr{13}). It may indicate to the auditor how much the company is by itself incentivized to take action.

\textit{Planned Solution: Use Case Fit}:
The planned solution is investigated for its fit to the use case. That involves the ML model itself but also its application. As a first step, it should be assessed if there is a viable alternative to the ML model in the form of a deterministic solution or a rule-based set that should be used instead (\intvwr{5}, \intvwr{12}). Such solutions tend to be less affected by biases (\intvwr{4}, \intvwr{5}). It can also be the case that a manual process should be preferred based, for instance, on ethical considerations (\intvwr{12}).
If an ML-based solution is chosen, model size and the amount of data flowing into a model should also be assessed (\intvwr{5}). The bigger a model gets, the larger is the potential and complexity regarding biases (\intvwr{3}). Smaller models usually also offer the advantage of better explainability (\intvwr{4}), which is a topic discussed with the concept \enquote{Model Reporting and Transparency} in the chapter \enquote{Criteria for Governance}. However, there is typically a trade-off in other KPIs like performance (\intvwr{4}, \intvwr{5}).
Moreover, the chosen objective should be assessed for certification. It should not represent a loosely connected and just correlated proxy for the business problem at hand but rather represent it most accurately (\intvwr{12}). Often, correlated information on ethnicity is used for predicting an objectively unrelated variable leading to harmful results, for instance, for the black community (\intvwr{12}). To check for that issue, the company should state its business goal so the certifying institution can assess the chosen objective’s suitability (\intvwr{12}). A mismatch should be easily detectable by an auditor (\intvwr{12}).
\subsubsection {Fairness Targets in Requirements}
Another finding is that fairness should be embedded in the requirements set for the NLP project (\intvwr{2}, \intvwr{4}, \intvwr{9}). Fairness-related criteria should be kept in an abstract form to be applicable to all projects (\intvwr{2}). When defining tasks, these requirements should help to give direction in the form of a checklist of where to focus and where problems may occur, hence introducing a reflection process (\intvwr{2}). A requirements catalog should direct the development and improvement of the system towards systematically including fairness considerations by enforcing specific steps for fairness to be considered (\intvwr{9}). It embodies the company’s vision regarding fairness and its commitment to it (\intvwr{2}, \intvwr{9}). Such a requirements catalog or checklist may be assessed in an audit (\intvwr{9}).

\subsection{Data-related Criteria} 
Most interviewees agree on the relevance of data-related criteria for certification of fairness (\intvwr{1}, \intvwr{4}, \intvwr{5}, \intvwr{6}, \intvwr{8}, \intvwr{9}, \intvwr{10}, \intvwr{13}, \intvwr{14}). One general aspect that should be considered regarding data is, that all processes leading to the data that the model eventually uses, should be reproducible for an audit (\intvwr{4}). Apart from the data assessment itself, it is also discussed how data should be managed and how it should be handled and transformed in processes over the lifecycle like annotation, preprocessing, evaluation, and continuous improvement.
\subsubsection {Data Assessment}

\begin{figure*}[htbp]
   \centering
   \includegraphics[width=1\textwidth]{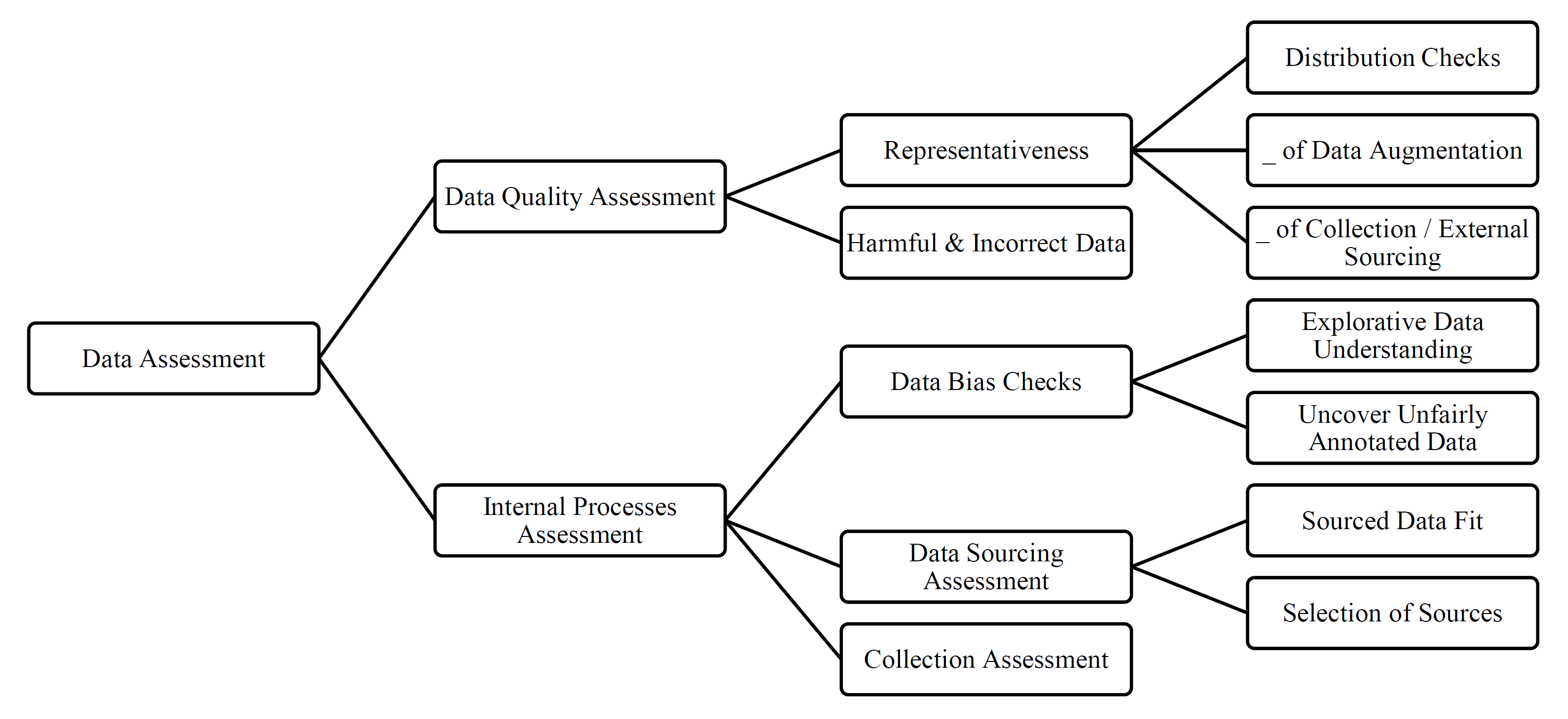}   
   \caption{Criteria relevant to \enquote{Data Assessment} hierarchically mapped}
   \label{fig:7}
\end{figure*}

Data assessment focuses on judging the suitability of data for training and retraining models that operate fairly. The data set on which it is trained can be seen as the major fairness issue (\intvwr{1}, \intvwr{2}, \intvwr{5}, \intvwr{6}, \intvwr{7}, \intvwr{10}), which gives reason for a thorough investigation. It needs to be assessed regarding its fairness, specifically regarding the use case it is intended to be used for (\intvwr{1}). The data's specific impact on the overall system needs to be considered (\intvwr{5}, \intvwr{7}). Subject matter experts may be required to assess data regarding fairness in an audit (\intvwr{12}). However, it needs to be considered that data may be scarce for specific use cases, leaving little room for improvements for the audited company (\intvwr{6}). As another limitation, interviewee five mentions that some information may also not be available for assessment in certain cases (\intvwr{5}). Data assessment entails an assessment of the data quality by the auditor but also an assessment of company processes that aim at avoiding data affecting fairness negatively. All relevant concepts for data assessment are portrayed in Figure \ref{fig:7}.

\textit{Data Quality Assessment}:
Data quality is a concept that covers to what extent data captures reality completely and in an undistorted way while keeping suitability for the use case in mind \cite{pushkarna2022data}. Other facets of data quality are not discussed here. Representativeness (\intvwr{1}, \intvwr{2}, \intvwr{4}, \intvwr{6}, \intvwr{7}, \intvwr{8}, \intvwr{9}, \intvwr{10}, \intvwr{11}, \intvwr{12}, \intvwr{13}) and identifying harmful and incorrect data (\intvwr{2}, \intvwr{4}, \intvwr{8}, \intvwr{9}, \intvwr{10}, \intvwr{13}, \intvwr{14}) are mentioned for the data quality assessment. Representativeness captures data, including all the variables needed, to represent the diversity of users of the system (\intvwr{12}). Interviewee twelve mentions the prominence of lacking representativeness of certain ethnic groups, causing fairness issues for ML models in the past (\intvwr{12}). To assess the representativeness of data, it makes sense to introduce distribution checks (\intvwr{1}, \intvwr{4}, \intvwr{8}, \intvwr{9}, \intvwr{13}). Distributions can be aimed at being representative of overall society (\intvwr{10}, \intvwr{12}) or the system’s users (\intvwr{2}, \intvwr{6}, \intvwr{7}, \intvwr{11}). Lacking representation in distributions can either happen by the exclusion of certain groups (\intvwr{6}, \intvwr{7}) or by skewed representation (\intvwr{1}, \intvwr{4}, \intvwr{6}, \intvwr{13}) and should be explained (\intvwr{6}). Representativeness is assessed for different attributes or target variables represented in data distributions (\intvwr{4}, \intvwr{13}). For assessment, statistic screening (\intvwr{12}) and frequency-based screening (\intvwr{4}, \intvwr{10}) may be utilized. The latter could also be described as a NER-based representations screening (\intvwr{10}). Words indicative of a class for a sensitive attribute and related words are automatically tracked via their semantic vector representation, which mirrors what NER does (\intvwr{10}). Their frequencies are compared to the representative real-world ratio. A comparison to a reference dataset regarding frequencies and semantic representation may also be utilized (\intvwr{10}). Checking label distributions in the annotation process may uncover some underlying data issues (\intvwr{6}). One particular factor that may impact representativeness after data acquisition is preprocessing. Representativeness may be negatively affected by data cleaning in the form of the removal of outliers (\intvwr{10}, \intvwr{12}) or by sampling (\intvwr{4}). That makes assessing distributions before and after preprocessing necessary (\intvwr{12}).
The sourcing and collection of data need some assessment by the certifying institution regarding representation issues. It should be checked whether the data distribution represents the use case (\intvwr{2}, \intvwr{6}, \intvwr{8}). As data is sometimes procured from abroad and collected there in an entirely different societal background, there may be issues with its representativeness for the use case (\intvwr{6}, \intvwr{8}). It could be checked in an assessment whether the application context of data meets its distributions. Similar issues occur when poorly fitting open datasets are used for first model iterations (\intvwr{2}). Induced selection biases can be found by inspecting the underlying data collection process (\intvwr{6}, \intvwr{12}). The context of time (\intvwr{12}), process characteristics, education of subjects, and location of the collection could be considered and questioned regarding possible adverse downstream effects (\intvwr{6}).
Data augmentation also needs to be mentioned, considering the representativeness of data (\intvwr{11}, \intvwr{13}). It can be assessed whether the baseline set for augmentation captures diverse enough data to represent the system’s users (\intvwr{11}). Because then there likely will not be a negative impact on representativeness after augmentation (\intvwr{11}). A company could even utilize data augmentation to improve the representativeness of data (\intvwr{13}). That may be done by augmenting larger quantities for the underrepresented group. However, this comes with the limitation of a loss in label accuracy (\intvwr{13}). The approach to data augmentation and its downstream impacts may be relevant to check in an audit.
Online learning utilization may enable a system to adjust automatically toward better representation (\intvwr{9}). It captures systems that automatically go through continuous improvement cycles by automated capturing of data for retraining (\intvwr{9}). When representation issues can be identified, the system could be set up to counter-steer automatically (\intvwr{9}).
Harmful or incorrect data negatively impacts data quality, and systems should be checked for it (\intvwr{2}, \intvwr{4}, \intvwr{10}, \intvwr{13}, \intvwr{14}). Interviewee 13 mentions harmful contents in baseline data as an issue that should be checked, particularly for generative models (\intvwr{13}). Misinformation in data is also an issue that should be investigated (\intvwr{8}, \intvwr{13}). However, in some cases, there is a grey area where it is difficult to judge to which extent data accurately represents the true nature of facts (\intvwr{8}). The last factor that should be assessed regarding harmful data is word embeddings which become relevant in transfer learning (\intvwr{2}, \intvwr{4}, \intvwr{9}, \intvwr{14}). The semantic representation of words in vector space needs to be checked for unwanted associations like gender with occupation (\intvwr{2}, \intvwr{4}, \intvwr{14}). Therefore, distances between embeddings in vector space should be checked before training a specific model (\intvwr{4}, \intvwr{14}). A reference data set may again be established to set an expectation for vector space representation (\intvwr{10}). Embeddings do have a downstream impact. Companies can handle the impacts of vector space representation by utilizing readjustable models (\intvwr{4}) or in their decision-making by compensating for their impact (\intvwr{14}). Such coping mechanisms may also be an aspect to assess for certification.

\textit{Internal Processes Assessment}:
Apart from investigating the data itself, a company's processes for sourcing, collecting, and checking it for biases should be assessed. Such mechanisms may be essential to consider as they can indicate how much stability of fairness in data over time can be expected. The data sourcing assessment investigates how the company procures its data and what considerations are made to ensure sourcing quality data for building a fair system (\intvwr{13}). The selection process for sources is a central consideration here (\intvwr{1}, \intvwr{2}, \intvwr{4}, \intvwr{8}, \intvwr{12}, \intvwr{13}). There may, for instance, be quality indicators like reviews or reputation of a data set of interest (\intvwr{1}), ratings for articles or scientific journals utilized (\intvwr{13}), or, in the case of social media data, for instance, an up-vote to down-vote ratio (\intvwr{4}). Social media, however, needs careful consideration regarding the intended use case as it tends to be very opinionated and often biased (\intvwr{4}). Sources renowned for extreme perspectives, which tend to be biased, should be avoided (\intvwr{4}, \intvwr{8}). When sourcing externally, it should be checked for transparency about the data collection process (\intvwr{1}, \intvwr{12}). Sources used should be documented (\intvwr{4}), and the sourced data should be reviewed regarding use case fit (\intvwr{2}, \intvwr{4}, \intvwr{5}, \intvwr{6}, \intvwr{13}).
The data collection assessment checks for suitable processes to collect data (\intvwr{12}, \intvwr{14}). The collection of data by the company itself, in contrast to data sourcing, gives the company complete control over its data acquisition (\intvwr{12}). Guidelines for the use case should be set for the collection process to avoid bias and may be reviewed for certification (\intvwr{3}). Data is sometimes synthetic and, in the case of chatbots, sometimes a result of brainstorming different ways of expression for responses in the development team (\intvwr{2}). That is why special guidelines may be required to avoid inducing bias (\intvwr{2}).
Assessing forms of bias checks a company performs may help assess processes enforcing fairness (\intvwr{1}, \intvwr{6}) at the cost of being resource intensive (\intvwr{2}). Human involvement should be checked, as it makes sense to pursue a semi-automated approach in this case (\intvwr{2}). That is because a human has a more refined sense of detecting relevant biases or fairness issues than a machine (\intvwr{2}). Hence, it should be checked whether a human assesses a data sample for potential fairness issues (\intvwr{2}, \intvwr{10}). The company should establish processes for an explorative data understanding of the dataset, helping to assess its quality and contents (\intvwr{1}, \intvwr{4}, \intvwr{5}, \intvwr{10}, \intvwr{11}). The contents, like ways of expression (\intvwr{5}, \intvwr{9}), contained entities (\intvwr{5}), and personalized data (\intvwr{5}), already indicate potential biases in data towards which a system needs to be robust. Even just the language that is used may indicate whether a system is susceptible to gender bias (\intvwr{13}). Moreover, data contents should also be internally checked for diversity and inclusiveness (\intvwr{4}, \intvwr{7}, \intvwr{9}, \intvwr{11}), potential stereotypes represented, and harmful content (\intvwr{1}, \intvwr{8}). A company should investigate all of this, considering metadata describing data sources to get the context that may be relevant for identifying issues (\intvwr{11}). Data bias checks should also help the company uncover unfairly annotated data (\intvwr{6}, \intvwr{12}). That can be done by comparing incorrectly labeled data between groups upon review (\intvwr{12}). If such unfairness is discovered, a company should investigate why it happens (\intvwr{6}). In some cases, a company may uncover bias in data, but there may be no suitable mitigation strategies. That should lead to documentation of the issue for mitigation in downstream processes (\intvwr{3}).
\subsubsection {Annotation}
The annotation process of data can be considered an important factor for the fairness of a system based on supervised learning (\intvwr{4}, \intvwr{6}). When investigating the annotation process of a company in an audit, it may be challenging to assess the quality of annotations in some cases as there may not be a ground truth that can be utilized for evaluation (\intvwr{12}). Moreover, there is a high degree of subjectivity to annotation (\intvwr{5}). If there is an objective ground truth, however, one may assume that a correct label is a fair label and evaluate fairness by the correctness of labels (\intvwr{12}). To certify annotation, one may consider processes established in a company for annotating data. Three major concepts are identified for company processes.
First, fairness-relevant aspects in the data should be annotated in the data (\intvwr{1}, \intvwr{2}, \intvwr{11}). That is a crucial step for being able to evaluate the fairness of the system for an audit or continuous improvement (\intvwr{1}, \intvwr{2}). Another way this benefits the development of fair systems is that it facilitates purposeful sampling allowing for fair representation of certain groups in data (\intvwr{2}). The annotation process of fairness-relevant aspects may either be manually labeled or automatically tagged (\intvwr{2}). However, one drawback to such annotations is also mentioned with its possible negative performance impact (\intvwr{1}).
Second, there is a tradeoff between labor cost and time and the quality of annotations to handle (\intvwr{2}, \intvwr{11}). To a certain extent, saving cost and time will sacrifice quality and hence fairness of annotations. Interviewee four proposes an approach that may be requested of companies as an option to pursue a more reflected annotation strategy regarding label quality. This tradeoff can be managed appropriately by including filters defining what needs to be annotated manually (\intvwr{4}).
Third, there should be a set of measures implemented by the company that causes a minimization of fairness risks in the annotation process (\intvwr{4}, \intvwr{5}, \intvwr{7}, \intvwr{10}, \intvwr{12}, \intvwr{13}). Like every human, human annotators have biases that motivate annotation processes so that their biases are at least less reflected in their annotations of data (\intvwr{7}, \intvwr{11}). While annotating data, one may require the annotator to see similar annotated examples to the respective one that should be annotated (\intvwr{7}), facilitating a more objective decision. One may also require the process to be designed so that sensitive information stays hidden and cannot bias the annotators’ decisions (\intvwr{7}). An aspect frequently mentioned is involving multiple annotators in labeling a data point (\intvwr{4}, \intvwr{5}, \intvwr{7}, \intvwr{10}, \intvwr{12}, \intvwr{13}). Interviewee four suggests that it should be case-dependent to either require multiple annotators or not (\intvwr{4}). Particularly for topics that are very opinionated and perceived differently by every individual, like politics, multiple annotators should be involved to reduce biases (\intvwr{4}). The aggregation of multiple annotators’ opinions on the correct label can either be achieved by a majority vote between annotators (\intvwr{5}, \intvwr{12}) or by an Inter-Annotator Agreement (\intvwr{4}, \intvwr{7}, \intvwr{10}, \intvwr{13}). An Inter-Annotator Agreement score quantifies how similarly multiple annotators annotate the same pieces of text (\intvwr{4}, \intvwr{10}). Particularly referring to topics that tend to be opinionated, a very high score represents a high agreement which may indicate that annotators are in a bubble (\intvwr{10}). It should raise organizational questions which are discussed in \enquote{Criteria for Governance}. However, in the case of developing generative models, finding a suitable metric as a baseline for such an Inter-Annotator Agreement may be difficult (\intvwr{5}). Another integral approach the audited company should pursue is utilizing annotation guidelines (\intvwr{4}, \intvwr{6}, \intvwr{7}). These guidelines should be documented so they can be audited and as objective as possible (\intvwr{4}). A beneficial consequence of introducing annotation guidelines besides fairness is more consistency and hence higher quality of annotation (\intvwr{6}).
\subsubsection {Preprocessing}
Preprocessing includes all steps for transforming, cleaning, selecting, and expanding data after annotation for training and evaluation. When auditing this step in the lifecycle, it must be considered that its scope may vary strongly between use cases (\intvwr{3}, \intvwr{11}). That makes it problematic to require certain steps for all use cases. Instead, the steps taken should be documented and assessed regarding fairness impact (\intvwr{4}). One concept mentioned several times is that the steps in preprocessing should be checked to determine whether they induce bias or counteract measures to improve fairness from previous lifecycle steps (\intvwr{10}, \intvwr{11}, \intvwr{12}). The same criteria established to be met by the data before preprocessing should also be met after preprocessing (\intvwr{12}).
Robustness-ensuring mappings are a way to transform data by mapping it to a standard form (\intvwr{2}). They may be helpful to pursue in some use cases as they can remove usability hurdles (\intvwr{2}). They can be, for instance, utilized to enable people with migration backgrounds to achieve a similar performance compared to native speakers of a language. For instance, articles can be mapped to a generic domain, making typical mistakes irrelevant (\intvwr{2}).
Anonymization may be a very related approach because it aims to drop the sensitive attribute in the data (\intvwr{3}, \intvwr{5}, \intvwr{7}, \intvwr{14}). The issue with such an anonymization approach is that dropping the sensitive attribute makes it challenging to assess the fairness of a model at a later stage, as the information from the sensitive attribute is lost (\intvwr{7}). Because of that, anonymization counteracts the annotation of fairness-relevant attributes as an earlier step in the lifecycle. Another problem with anonymization is that it does not necessarily solve the problem of biased data, as there usually are indirect encodings through correlations that can materialize as second-order effects (\intvwr{8}). In some use cases, anonymization may not be suited as it may undermine the solution’s functionality because of a negative performance impact (\intvwr{5}, \intvwr{10}). The same applies to differential privacy, which is an approach for comparing the effect of including the sensitive attribute and dropping it (\intvwr{5}, \intvwr{7}). Nevertheless, in some cases, it may be viable and should be considered by the company (\intvwr{5}, \intvwr{7}).
An important consideration for preprocessing that may be assessed is the filtering and selection of data (\intvwr{3}, \intvwr{4}, \intvwr{13}). Filtering and selection aim to improve the data quality for use via filtering mechanisms and ensure appropriate data splitting. Filtering may target removing harmful content like obscene language (\intvwr{13}) or low-quality or biased data (\intvwr{3}, \intvwr{4}). However, it may be challenging for some topics to identify biased data clearly (\intvwr{13}). Moreover, defining fair filters objectively and calibrating how restrictive they should be poses a challenge (\intvwr{4}). Particularly the use of social media data raises the question of controllability for such a process (\intvwr{13}).
When selecting data, unbiased and balanced data sampling is essential (\intvwr{1}, \intvwr{4}, \intvwr{7}, \intvwr{8}, \intvwr{10}, \intvwr{12}, \intvwr{13}). Depending on how data is selected, a biased selection could be made, which is to be avoided (\intvwr{1}, \intvwr{7}, \intvwr{10}, \intvwr{12}). When splitting the data into training data and data for evaluation, it should be assessed whether the distributions of training, evaluation, and overall data roughly match (\intvwr{10}, \intvwr{13}). As mentioned, that helps avoid bias and ensures the integrity of evaluation results (\intvwr{10}). It was also mentioned that the data splitting might consider underrepresented minorities by representing them in the data samples equally compared to other instances that are better represented (\intvwr{10}).
\subsubsection {Data for Evaluation}
In evaluation, some aspects and concepts regarding data must be considered. Functional testing of a system is input-based testing assessing the outputs and hence views the system from an external perspective suited for an audit (\intvwr{10}). The procedure is discussed in the chapter \enquote{Modeling \& Evaluation Criteria}. Test data is targeted at unveiling biases present in the model (\intvwr{3}). It should represent the diversity of actual usage of the system and the challenges that come with it (\intvwr{2}, \intvwr{11}). Moreover, it should suit the assessment regarding the specific fairness definition imposed by the audit (\intvwr{9}).
A relevant consideration is the sourcing of data sets utilized for evaluation. It could be acquired by accumulating in-use data of the system over time (\intvwr{12}). Another way of acquiring test data is going by previous precedents of fairness issues and utilizing relevant data regarding these (\intvwr{13}). The last approach to getting data is artificially generating it, making a targeted evaluation of fairness easier as it can be designed to reveal biases (\intvwr{2}, \intvwr{6}). Targeted segmentation for sampling data of inputs can help assess specific biases (\intvwr{7}). However, these data sets may also be biased and only provide partial insight into fairness issues (\intvwr{8}).
Another vital consideration for functional testing is utilizing fairness invariance tests (\intvwr{2}, \intvwr{3}). Such tests utilize targeted replacements in inputs to assess the model’s invariance to variation in sensitive attributes’ values (\intvwr{2}). These replacements, as a form of augmentation, are usually generated by multiple language models (\intvwr{2}) and involve approaches like forward and backward translation (\intvwr{2}, \intvwr{3}). Drawbacks to this approach are mentioned with the complexity of such augmentations in NLP (\intvwr{3}) and potential model biases present in the models utilized for generating replacements (\intvwr{3}). Nonetheless, interviewee two sees robustness invariance tests as an integral part of certifying the fairness of a system (\intvwr{2}). It is suggested to propose a range of measures and a minimum standard for creating such augmentations for fairness invariance tests (\intvwr{2}).
\subsubsection {Data for Continuous Improvement}
Data-related criteria in continuous improvement target data considerations for maintaining a fair system over the time of its operation. That involves considering monitoring practices as well as feedback loop practices. Continuous improvement is affected regarding data by two challenges: maintaining data diversity (\intvwr{9}) and extending continuously relevant testing data (\intvwr{9}).
As criteria for monitoring, test sets (\intvwr{2}, \intvwr{3}, \intvwr{6}, \intvwr{8}, \intvwr{9}), drift monitoring (\intvwr{2}, \intvwr{3}, \intvwr{4}, \intvwr{5}, \intvwr{6}, \intvwr{7}, \intvwr{13}), as well as a request assessment for underrepresented groups (\intvwr{2}) are mentioned as practices to be implemented by the audited company. Test sets in continuous improvement are the same data-wise compared to test sets in evaluation, but they are used continuously to assess whether fairness can be maintained or whether certain issues have occurred (\intvwr{2}, \intvwr{3}, \intvwr{6}, \intvwr{8}, \intvwr{9}).
Drift monitoring is described either as data-centric (\intvwr{2}, \intvwr{4}, \intvwr{5}, \intvwr{7}, \intvwr{13}) or as model-centric (\intvwr{2}, \intvwr{4}, \intvwr{5}) in its approach. It may reveal the necessity to retrain a system to stay fair and performant over time (\intvwr{2}, \intvwr{3}, \intvwr{4}, \intvwr{5}, \intvwr{6}, \intvwr{13}). A model-centric approach considers model behavior like changing performance or confidence in predictions (\intvwr{2}, \intvwr{5}) and explainability (\intvwr{2}, \intvwr{4}) of the latter, in the form of explanation scores or some other form of quality assessment for the explanation, in detecting drifts. In a data-centric approach, the variation of incoming data for the model to handle compared to the original training data is assessed (\intvwr{2}, \intvwr{4}, \intvwr{5}, \intvwr{7}, \intvwr{13}). Continuously appearing outliers may be an indication (\intvwr{2}). It may be helpful to trace back to the kind of requests causing the drift to understand the underlying issue (\intvwr{2}). The reaction strategies to an occurring drift may be interesting for an audit (\intvwr{3}). When users change, assessing them for semantic overrepresentation, which may cause an unfavorable data drift, may also be relevant (\intvwr{2}). The frequency at which such drifts may occur is dependent on the dynamics of the environment in which the application is operational (\intvwr{2}, \intvwr{5}). That should impact the assessment of the calibration of such a drift monitoring system. An assessment should also consider that drift detection may lack a measurable foundation for some use cases (\intvwr{3}). Hence, it should only be required for changing environments and if it is measurable. The consequences of drift monitoring may materialize in improved fairness and performance, making it compatible with business goals (\intvwr{2}, \intvwr{3}).
An assessment of requests for underrepresented groups focuses on better understanding marginalized groups’ requests (\intvwr{2}). That helps uncover potential issues to counteract the negative impact of underrepresentation in data they face (\intvwr{2}). That may be performed by checking semantic differences in the expression of marginalized groups compared to better-represented groups (\intvwr{2}). Substantial deviations may indicate potential issues in system performance later (\intvwr{2}).
Feedback loop practices regarding data need to consider how in-use data and user-made corrections are handled. One condition that needs to be met by both is user privacy (\intvwr{7}, \intvwr{9}, \intvwr{10}). That may target a confidential treatment of user feedback (\intvwr{7}) as well as removing sensitive attributes from that feedback (\intvwr{10}) and ensuring data protection and data security when saving data (\intvwr{9}). In-use data is gathered by regular operation of the system without any user interaction targeted directly at generating feedback and should also be subject to assessment (\intvwr{5}, \intvwr{9}). One form of in-use data that can be utilized is prediction queries dependent on confidence in the prediction, either taken for retraining future models in case of low confidence or for usage as test data otherwise (\intvwr{9}). That works in favor of underrepresented groups as their queries are more likely to be predicted with low confidence and hence taken for model improvement. Metadata may be essential to understand fairness-relevant aspects in such a case (\intvwr{6}, \intvwr{9}). However, if metadata is unavailable, the context of the query stays hidden, and it may be challenging to target model improvement toward fairness (\intvwr{9}). Another form of in-use data can be gathered by indirectly measuring user discontent by analyzing behavior when the user interacts with the application (\intvwr{5}). That may indicate a system’s impact on its users, and after passing a certain threshold, it can be estimated to be unfair, requiring intervention (\intvwr{5}). The indirect form of measurement offers the advantage of gathering data without influencing or biasing the user in their feedback statement, which may happen when directly asking for it (\intvwr{5}). A dedicated team or a quality assessor should check usage data regularly for such aspects (\intvwr{5}).
If users can make immediate corrections or give feedback which may be taken to improve the system, particularly the quality of feedback data needs attention as feedback may be biased or incorrect (\intvwr{4}). If the user is facilitated to correct an annotation, clear guidelines should be established by the company and communicated to the user (\intvwr{4}). Hence, data is consistent with the originally annotated data, and incorrect annotations may be prevented. The latter should also be enforced by a controlling instance that has to review and accept user annotations (\intvwr{4}). Moreover, there should be some form of filter rejecting harmful, like for instance stereotyped, feedback data generated by the user from being utilized for model retraining (\intvwr{10}). There are industry examples where such data has caused a system to become heavily biased (\intvwr{9}, \intvwr{10}).
\subsubsection {Data Storage}
Regarding data storage which covers how data is stored in development and operations and cached in operations, two concepts are introduced. First, the storage of sensitive attributes should be avoided in a raw form (\intvwr{10}). That has gained relevance due to the trend towards the ELT (Extract Load Transform) paradigm, as there is an interest in utilizing the raw and not the transformed data (\intvwr{10}). Avoiding saving sensitive attributes raw can also be partially motivated by compliance with GDPR, as personally identifiable information should not be saved (\intvwr{10}). This focus on privacy highlights again that fairness is interrelated with other RAI topics.
Second, interviewee one mentions caching of previous predictions of the system as a potential issue (\intvwr{1}). In case of unfair previous predictions, this necessitates clearing the cash to restore fairness of the system (\intvwr{1})).

\subsection{Modeling \& Evaluation Criteria} 
In this chapter, fairness criteria for building NLP models and evaluating them and the system they are embedded in are explained.
\subsubsection {Modeling}
Modeling, particularly model selection, is perceived as less critical for certifying fairness compared to criteria targeting other lifecycle steps (\intvwr{1}, \intvwr{5}, \intvwr{10}). A potential issue mentioned regarding the utilization of pretrained models is a lacking holistic understanding of it (\intvwr{3}), making it hard to assess and enforce fairness for training the custom model. That leads interviewee three to suggest only permitting the usage of, to some degree, trustworthy and documented baseline models for certification (\intvwr{3}). That could solve the existing business process and documentation conflict in case a poorly documented model outperforms a well-documented one (\intvwr{3}). This would also incentivize providers of performant models to document them appropriately.
When developing and training models, fairness can be rooted in the model architecture (\intvwr{2}, \intvwr{4}, \intvwr{5}, \intvwr{6}, \intvwr{7}, \intvwr{10}, \intvwr{14}). For doing so, interviews reveal four potential strategies. The first introduces a fairness classifier built onto the model used (\intvwr{5}). For implementing something like this, however, a labeling of fairness for all annotations used in the original model would be required, which would be very resource-consuming and impractical (\intvwr{5}). The second approach embeds fairness into the objectives optimized by the model (\intvwr{4}, \intvwr{5}, \intvwr{7}). That may be done by performing a hyper-parameter optimization of subjective fairness, even though it would be time and resource-consuming (\intvwr{5}). Another way would be to calculate the loss function, which is to be optimized only on non-sensitive attributes or to penalize using the latter (\intvwr{10}). Either way, the question needs to be asked whether the intended model is suitable for such approaches in practice (\intvwr{7}). The third approach aims at explicitly calibrating a model to avoid certain biases or, in a way, hard-code fairness into the system (\intvwr{4}, \intvwr{5}, \intvwr{6}, \intvwr{14}). That may go beyond the NLP systems prediction and involve rules for decision-making based on the prediction (\intvwr{6}, \intvwr{14}). As a fourth approach, interviewee two suggests introducing constraints for model controllability (\intvwr{2}). Language generation via a prompt-based model is mentioned as an example where there should be the possibility to avoid biases by setting certain parameters influencing constraints for the model (\intvwr{2}). Otherwise, the model would mirror potential biases in baseline data (\intvwr{2}). All of the mentioned approaches should be assessed regarding their impact on the performance and fairness of the system (\intvwr{5}).
\subsubsection {Evaluation}
Evaluation is essential for assessing a model's fairness (\intvwr{3}, \intvwr{7}, \intvwr{9}). It involves testing the model or the system it is embedded in regarding various targets or metrics. Reproducibility must be given for company internal evaluation processes so they can be assessed in an audit (\intvwr{8}). A challenging factor may be the definition of fairness which is the foundation for evaluating fairness (\intvwr{8}, \intvwr{12}). As the definition of fairness differs between the auditor and all affected stakeholders (\intvwr{12}), a hierarchical approach to evaluation focused on the system’s stakeholders may make sense as a baseline for evaluation (\intvwr{8}). Another challenging factor is to be found in the context of data as a prediction foundation. Sometimes truthfulness or accuracy may be questionable but hard to verify, posing a major issue for evaluation (\intvwr{8}). In open-domain systems like question-answering systems or recommender systems delivering multiple suggestions, the complexity of human language becomes a challenge for fairness evaluation (\intvwr{2}, \intvwr{8}).
The auditor may assess all criteria in this chapter directly as system tests. However, they should also be embedded in the company’s internal evaluation processes, which can be audited. Interviewees name two major concepts that drive evaluation. Functional testing criteria (\intvwr{1}, \intvwr{2}, \intvwr{3}, \intvwr{4}, \intvwr{5}, \intvwr{6}, \intvwr{7}, \intvwr{8}, \intvwr{9}, \intvwr{11}, \intvwr{12}, \intvwr{13}, \intvwr{14}) and metrics (\intvwr{2}, \intvwr{4}, \intvwr{7}, \intvwr{8}, \intvwr{9}, \intvwr{10}, \intvwr{14}) are defining factors identified for assessing fairness in evaluation.

\textit{Functional Testing Criteria}:
Functional testing criteria take an outside perspective on the system’s outputs given some specific inputs. An outside perspective on the application is taken, and no deep insights into the model architecture are utilized here (\intvwr{9}). The outputs for a given set of inputs indicate model behavior and resulting fairness issues (\intvwr{1}, \intvwr{5}, \intvwr{6}, \intvwr{7}, \intvwr{9}, \intvwr{11}, \intvwr{12}, \intvwr{14}). Functional testing of the model can be considered an essential part of a certification process for fairness as it captures what the model predicts and what impacts its users (\intvwr{9}). In comparison, the already mentioned data is the model’s baseline, but with nonlinear relationships of a model, even seemingly fair data can result in biased models (\intvwr{8}, \intvwr{9}). Hence, functional testing criteria capture the actual fairness of a system better. For such an approach to be meaningful, a relevant scope of inputs is required regarding quantity (\intvwr{2}) and diversity of input data, representing a wide variety of user groups (\intvwr{9}, \intvwr{11}). That may be facilitated via a mainly automated approach to testing (\intvwr{2}, \intvwr{3}, \intvwr{4}, \intvwr{11}). 
Adversarial testing is mentioned for prompt-based generative language models (\intvwr{1}, \intvwr{4}, \intvwr{7}). In that case, it involves utilizing specific queries that suggest or invoke stereotypes is a practical approach for fairness testing (\intvwr{1}, \intvwr{4}, \intvwr{7}). If the assumption of the stereotype is met by the most probable words to be inferred by the generative model, a fairness issue is detected (\intvwr{1}, \intvwr{7}). These associations made by the model can be assessed in an audit (\intvwr{4}). Interviewee seven also points out the necessity of finding a metric for evaluating the fairness of a model going off a defined set of prompts (\intvwr{7}).
Another approach is utilizing specific test data sets (\intvwr{2}, \intvwr{3}, \intvwr{6}, \intvwr{8}, \intvwr{9}, \intvwr{11}, \intvwr{13}). The data utilized here has been discussed previously in the chapter \enquote{Data-related Criteria}. The mentioned data is used as model input with a specific expectation for the predictions’ quality (\intvwr{2}). It may make sense to analyze the distribution of predictions to assess their quality, which may reveal systemic issues (\intvwr{1}, \intvwr{11}, \intvwr{12}). Interviewee one highlights that biased distributions in data should be accepted as a reality (\intvwr{1}). However, a change in decision-relevant variables like qualifications in a recruitment context should also change the distribution of predictions for a specific group (\intvwr{1}).
Hence, relevant variables to the topic should be checked to be the decision drivers. The latter may be achieved utilizing explainability tools discussed in \enquote{Model reporting \& Transparency} in the \enquote{Criteria for Governance} chapter. Decision-irrelevant variables should, however, not influence model inference (\intvwr{2}). This can be checked by introducing robustness invariance testing (\intvwr{1}, \intvwr{2}, \intvwr{4}), which should be an essential part of a fairness audit (\intvwr{2}). It aims to assess how capable a system is of generalizing regarding a sensitive attribute. The latter may be, for instance, represented in expression difficulties by non-native speakers of a language (\intvwr{2}). Protected groups may be introduced, representing sensitive attributes that could become an issue in the use case (\intvwr{2}). By augmenting initial test data via replacements for protected groups or the addition of nonsense, the robustness of the model can be checked in the form of adversarial testing (\intvwr{2}, \intvwr{4}). For most use cases, a larger invariance between a protected group and the rest of the users results in improved fairness (\intvwr{2}, \intvwr{4}). However, there are exceptions, like the medical domain, where separation is essential and should be the focus (\intvwr{3}, \intvwr{13}, \intvwr{14}). Assessment should be performed automatically with the additional requirement of involving a human assessment of a subset of data as a plausibility check for the automated assessment (\intvwr{2}). This invariance testing is enabled by annotating fairness-relevant aspects like gender in data (\intvwr{2}).
Predictions of the model may also be required to be validated (\intvwr{1}, \intvwr{4}, \intvwr{6}). Simpler models and particularly rule-based sets have the advantage of being typically less affected by biases, and decisions based on them tend to be based on relevant factors (\intvwr{4}). Previously set expectations may also be used to validate a model’s prediction (\intvwr{1}, \intvwr{6}). A discrepancy between prediction and what is used for validation should be investigated (\intvwr{6}).
Interviewee five suggests affected stakeholders as a source of functional evaluation regarding the system’s fairness (\intvwr{5}). As a deciding factor to overcome individual differences in opinions on the fairness of the system, a majority vote or some other kind of threshold may be introduced (\intvwr{5}). The challenge of possible discrimination of minorities which such an approach could induce, can be addressed by sampling minorities representatively when selecting the composition of affected individuals responsible for evaluation (\intvwr{5}).
An audit assessment should also consider ethics and moral considerations relevant to the system (\intvwr{4}, \intvwr{11}). As fairness and ethics are closely related, there needs to be a discussion as to what extent fairness should represent normative values in a society (\intvwr{9}). A central question here is: Is it fair if a system results in morally questionable actions that affect all groups in the same way (\intvwr{9})? A good fairness understanding of the use case, which is discussed in \enquote{Process criteria} is required to resolve ambiguity.
The last criteria to consider for functional testing focus on the interaction of humans with computers and the involvement of humans to improve fairness. Humans sometimes tend to accept predictions of machines as accurate and tend to become complacent in reviewing system outputs before utilizing them for decision-making, which is called automation bias. This automation bias is to be avoided as it does not counteract biases in predictions from impacting decision-making based on them (\intvwr{11}). Hence, there should also be criteria for human decision-making that counteract automation bias (\intvwr{11}). Humans should generally be involved and in the loop when assessing system fairness (\intvwr{2}, \intvwr{7}, \intvwr{11}). Regarding the application of the model, there should be an assessment regarding the necessity of human involvement in decision-making. In specific, particularly critical applications, human involvement in decision-making may be crucial (\intvwr{12}). Many fairness issues are only apparent to humans and may currently not be automatically detectable (\intvwr{2}). Moreover, humans may give the automatic system a plausibility check (\intvwr{2}).

\textit{Metrics}:
Metrics aim to make the fairness of NLP models measurable. Metrics are of essential importance for a targeted, quantifiable and reproducible assessment of the fairness of an application (\intvwr{7}, \intvwr{8}, \intvwr{12}). Metrics based on a generalization of the model are named as well as metrics that follow an impact-based approach regarding fairness. Robustness and generalization may be seen as essential elements in assuring confidence in the model’s property of not generating unexpected, unfair results (\intvwr{2}, \intvwr{4}, \intvwr{8}). For the first, the expected calibration error is named as a metric keeping track of the robustness of a model (\intvwr{8}). Also, standard metrics commonly utilized in data science may already indicate the generalization of a model and hence may be assessed (\intvwr{4}).
The impact-based approach targets potential fairness issues for particular groups more directly (\intvwr{8}), which manifests in the composition of fairness metrics. A major point of discussion in such metrics is their level of specificity (\intvwr{3}, \intvwr{7}, \intvwr{14}). A specific metric for fairness will not easily transfer between use cases as there is yet to be a dynamic fairness metric capable of handling that variety (\intvwr{7}). Less specificity in the metric will lead to less relevant results in the specific use case (\intvwr{3}, \intvwr{14}). Interviewee eight hints at a trade-off between performance and fairness metrics that may need to be made by sacrificing some performance for the merit of the impact-based metric (\intvwr{8}). Also, some metrics should consider and capture diversity information as a context when assessing performance metrics (\intvwr{8}).
Another impact-based metric would be a performance disparity threshold that needs to be met (\intvwr{2}, \intvwr{7}, \intvwr{9}). The quality of service provided by the system should be evaluated regarding differences between groups, and a threshold should set a maximum acceptable deviation (\intvwr{7}, \intvwr{9}). That may be expanded beyond the mere performance of the system and might include factors like accessibility and usability of the system as well (\intvwr{2}, \intvwr{9}). The latter aims at removing barriers to system use (\intvwr{2}) and may be less quantifiable than just performance disparity.
Another approach would be introducing a metric that penalizes the model's use of sensitive attributes or unethical content. It would be used by the optimizer of the model when calculating loss (\intvwr{10}). By this kind of reinforcement, some major fairness issues that could occur may be ruled out (\intvwr{10}). For instance, in question answering, responses that contain insensitive or stereotyped content could be avoided (\intvwr{10}).
A wide variety of fairness metrics are also represented in different definitions of fairness available to choose from (\intvwr{8}, \intvwr{12}, \intvwr{14}). No matter what metric is used in the end, its impact on the technical system must be assessed, and its suitability and implications for the use case must be assessed (\intvwr{1}, \intvwr{8}).

\subsection{Operations Criteria} 
Bringing a system into operation and going through continuous improvement cycles to maintain its quality should be investigated in a fairness certification process. These steps in the lifecycle determine whether a system will maintain fairness over time (\intvwr{9}). Fairness can deteriorate in operations, for instance, at the cost of performance (\intvwr{2}). Interviewee six mentions that the certificate must maintain integrity over time (\intvwr{6}), which supports the relevance of system operations criteria for fairness certification. Operations need to build quick reaction strategies to counteract fairness issues that occur swiftly (\intvwr{1}, \intvwr{7}, \intvwr{11}). Standard pipelines for retraining and publishing models are not viable as they take too long (\intvwr{7}). Another consideration in operations is the resource distribution for models, which should not put fair models at a disadvantage regarding resource access compared to other models (\intvwr{10}).
When drastic changes are made or occur in a systems environment, there may be a requirement to renew the overall certification assessment or at least those assessments affected by the change (\intvwr{5}, \intvwr{7}). Such changes may come in various forms. A new version roll-out may be a reason (\intvwr{1}, \intvwr{7}, \intvwr{8}). There should also be detailed documentation throughout model versions to ensure traceability when a fairness issue occurs (\intvwr{1}). Adaptions in data (\intvwr{1}, \intvwr{7}, \intvwr{9}), changing stakeholders (\intvwr{7}), or a change in the system's application (\intvwr{8}) may be further reasons for renewing affected assessment steps.
Deployment (\intvwr{2}, \intvwr{3}, \intvwr{6}), monitoring (\intvwr{1}, \intvwr{2}, \intvwr{3}, \intvwr{5}, \intvwr{7}, \intvwr{8}, \intvwr{9}, \intvwr{13}), and the feedback loop for continuous improvement (\intvwr{1}, \intvwr{2}, \intvwr{3}, \intvwr{4}, \intvwr{6}, \intvwr{7}, \intvwr{8}, \intvwr{9}, \intvwr{10}) are relevant to interviewees as the main criteria for operations,
\subsubsection {Deployment}
Criteria for deployment are mainly brought up in the context of providing different model sizes via pruning or knowledge distillation approaches (\intvwr{2}). First, it needs to be checked whether pruned or quantized models still fulfill fairness-related criteria established before to the same extent as their larger counterparts do (\intvwr{2}). It makes little sense to perform an intense assessment for the large model that the user does not utilize in practice and, for instance, miss out on functional assessment of the smaller model that a user may use more frequently (\intvwr{2}).
Smaller models tend to be more prone to underfitting issues and should be checked accordingly (\intvwr{3}). On the other hand, larger models may need more testing as they are opaque and tend to contain more complex and challenging to identify biases with more data being used for training them (\intvwr{3}, \intvwr{6}). When choosing a suitable model size, performance and fairness objectives should play a role (\intvwr{2}).
\subsubsection {Monitoring}

\begin{figure}[htbp]
   \centering
   \includegraphics[width=1\textwidth]{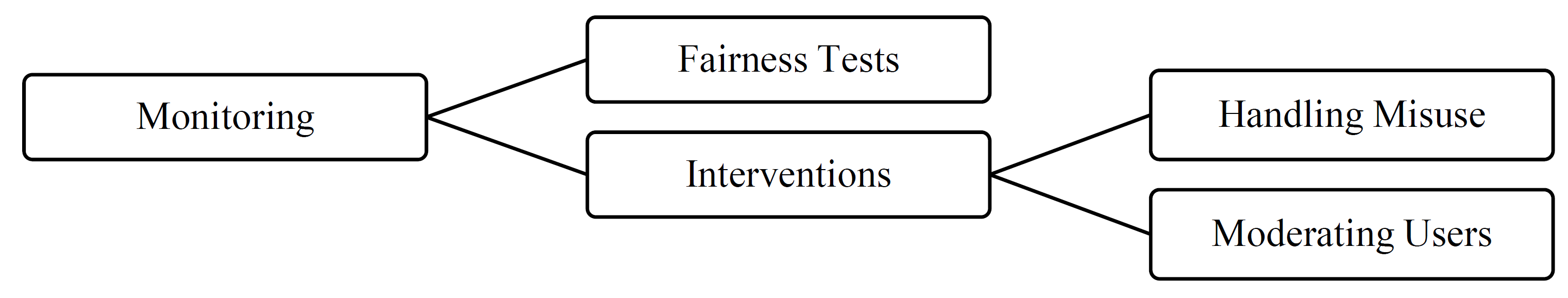}   
   \caption{Criteria relevant to \enquote{Monitoring} hierarchically mapped}
   \label{fig:8}
\end{figure}

Monitoring can be seen as a set of mechanisms that ensure fairness by only giving clearance for a new model version to be introduced or current versions to stay operational if fairness criteria can be assured (\intvwr{1}, \intvwr{5}, \intvwr{7}, \intvwr{9}, \intvwr{13}). It typically involves fairness testing and intervention mechanisms, as shown in Figure \ref{fig:8}.
Fairness tests have already been introduced under \enquote{Functional Testing Criteria} in the \enquote{Modeling \& Evaluation Criteria} chapter. Such an evaluation process based on fairness test sets should be extended to assess the model over time as it goes through continuous improvement cycles (\intvwr{1}, \intvwr{2}, \intvwr{3}, \intvwr{7}). The goal is to quantify the occurrence of fairness issues (\intvwr{3}) which may be valuable information for an audit. With every new model or change in data or parameters, the fairness dimension of the system may be affected (\intvwr{2}). Interviewee one mentions the importance of standards that can be systematically checked with every system retraining (\intvwr{1}). Such testing could be run at a very granular level assessing, for instance, confidence and explanation scores on a sentence level as a foundation for later judging fairness (\intvwr{2}). Passing scores or red flags should be established for fairness test sets to make the fairness testing impactful (\intvwr{2}, \intvwr{3}). Interviewees are divided on how automated such testing can be. While interviewees one and nine see potential in automated fairness testing, interviewee seven argues the process is not automatable as there is no dynamic fairness metric (\intvwr{1}, \intvwr{7}, \intvwr{9}). Depending on the automation of the testing, frequent testing is favorable (\intvwr{1}, \intvwr{3}). In large-scale organizations or in a highly dynamic context, where retraining models takes place multiple times a day, a fairness assessment with every retraining could cause a considerable overhead (\intvwr{3}). The company may execute this fairness testing process right after quality assessment to assess the fairness impact of changes in quality (\intvwr{9}).
In some cases, a system may be operated or operating in a potentially harmful way. To prevent the system from causing damage, intervention mechanisms are established. Particularly for transfer learning, there is the issue of a user or third party potentially misusing a system for unintended, unethical purposes like spreading misinformation, which needs to be avoided if possible (\intvwr{13}). Interviewee 13 mentions the curse of globalization in this context, as traceability of what is built with baseline models stays unknown to the provider (\intvwr{13}). Moreover, misuse has previously also caused issues with systems with automatic retraining on usage data (\intvwr{9}). Targeted, hostile user behavior in the form of interactions with the system can make it develop strikingly unfair tendencies (\intvwr{9}, \intvwr{10}). Manual controls and automated corrective mechanisms on a quantifiable foundation may be implemented to ensure system fairness over time (\intvwr{9}). Generally, a certification process could check whether the issue of misuse is addressed and whether potential countermeasures are in place.
Moderating users, on the other hand, can be an option to cope with user behavior that negatively affects a system's fairness (\intvwr{3}). Interviewee three mentions four shortcomings of moderating users of a system. Its practicality is questionable as a large volume of data needs to be handled, making it impossible to have human oversight (\intvwr{3}). That leads to utilizing models for moderation which may also be biased (\intvwr{3}). Moreover, moderation may just cause a problem shift by users utilizing competitors’ services, making moderation, not incentive compatible as it counteracts business goals (\intvwr{3}). In case a system develops extreme, unacceptable behavior, there also needs to be some form of kill switch, as running the system until the next update is rolled out may sometimes be unacceptable (\intvwr{7}).
\subsubsection {Feedback Loop}

\begin{figure*}[htbp]
   \centering
   \includegraphics[width=1\textwidth]{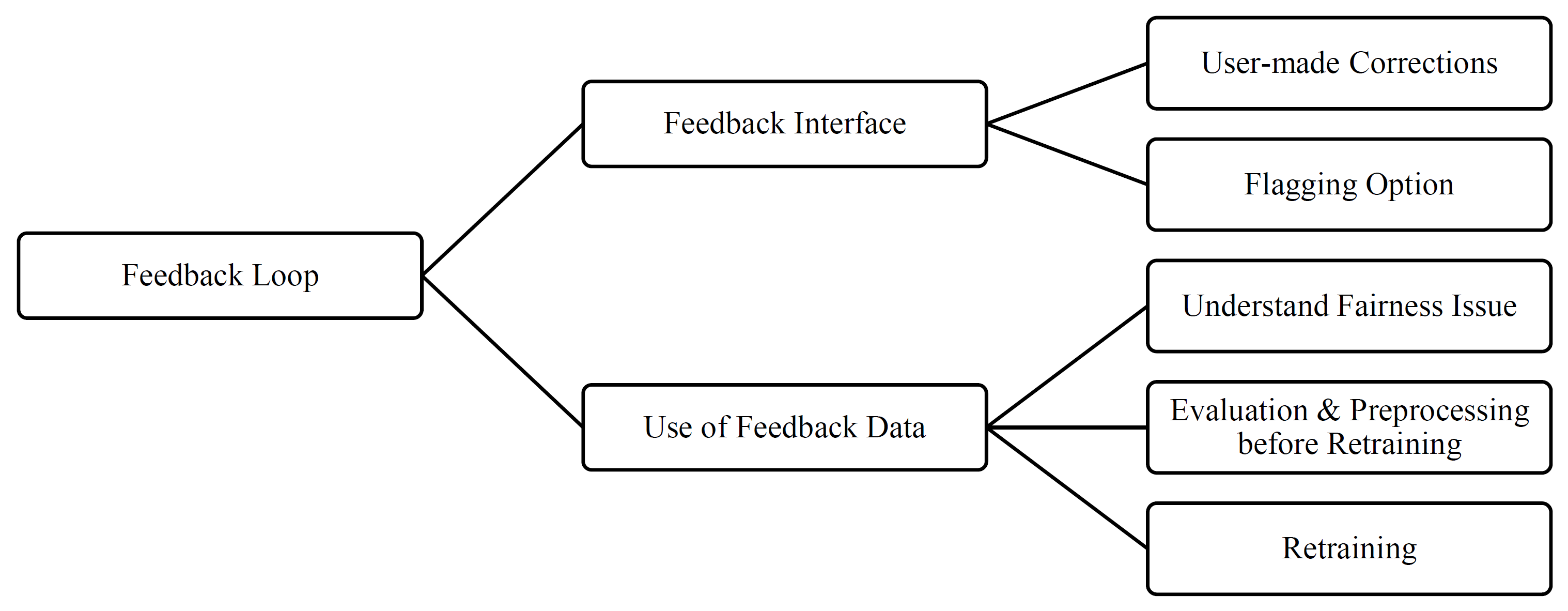}   
   \caption{Criteria relevant to the \enquote{Feedback Loop} hierarchically mapped}
   \label{fig:9}
\end{figure*}

Typically, a system is not designed to be static but rather dynamic, utilizing feedback data from its use for improvements (\intvwr{9}). That offers the advantages of outgrowing systemic biases (\intvwr{6}) and raising general model quality (\intvwr{2}), which aligns with business goals. A similar approach to active learning may be beneficial for creating an effective feedback loop for counteracting biases in the system (\intvwr{4}). By deducing systematic characteristics of biased cases and giving them as input back to users, their feedback can be more targeted (\intvwr{4}). To operationalize a system's feedback loop, an interface for the user to provide feedback or potentially even corrections for fairness issues is required. The acquired feedback data needs to be used appropriately to improve the system. The latter is displayed in Figure \ref{fig:9}.

\textit{Feedback Interface}:
The interface should allow users to interact with the system (\intvwr{1}, \intvwr{2}, \intvwr{3}, \intvwr{7}). A challenge such interfaces must overcome is engaging the user to utilize them (\intvwr{2}, \intvwr{4}). There may be a reluctance due to convenience (\intvwr{4}) or disengagement and disinterest in continuing to use the system if someone is offended by an occurring fairness issue (\intvwr{2}). An audit might consider the extent to which such interaction mechanisms are utilized to evaluate their effectiveness in exposing fairness issues so they can be addressed. In some cases, it may also be challenging to establish an interface in the first place as the user only indirectly encounters the model a few stages downstream in the process of an application (\intvwr{3}). So, an audit should consider how the model is embedded in a system. An organization may also have user-centric teams that extend feedback information acquired in a visual interface integrated into the application (\intvwr{7}). Those user-centric teams must closely collaborate with the organization's development team as their input may be required to identify previously unknown fairness issues, which can then be replicated and mitigated by the development team (\intvwr{7}). It may make sense to check for an interface for users to correct inference of a system (\intvwr{4}, \intvwr{8}). The number of corrections made by users may also be relevant information for a certification (\intvwr{4}). Interviewee 4 also mentions corrections that can be made for a model explanation as feedback on what users would base the prediction on (\intvwr{4}). The second form of interface targets reporting or flagging fairness issues encountered in use (\intvwr{1}, \intvwr{2}, \intvwr{3}, \intvwr{7}). The users should be able to express their discontent with the fairness issue that they should be able to describe (\intvwr{1}, \intvwr{2}, \intvwr{3}, \intvwr{7}). It should be checked whether such an interface is immediately accessible by a user (\intvwr{2}).

\textit{Use of Feedback Data}:
The issue with such interfaces is a tendency to shift accountability for fairness onto the system user, who may also be biased (\intvwr{6}). Hence, for certification, the use of feedback data and accountability that comes along with it should be checked. That covers correctly understanding the fairness issue at hand, evaluation and preprocessing of data that should flow into retraining a system, and mechanisms for deciding when to retrain the system. Understanding the fairness issue at hand involves the challenge of tracing the fairness issue back to the part of the system where it originates (\intvwr{1}, \intvwr{8}). Mechanisms for understanding the cause of the fairness issue should be implemented (\intvwr{1}, \intvwr{2}, \intvwr{3}, \intvwr{7}). Assessing the expectation of the user and his or her explanation of unfairness for the prediction may be a first step in understanding the issue (\intvwr{1}). Tracking who specifically reports fairness issues may hint, for instance, at a marginalized group that may have been neglected in development (\intvwr{2}). Depending on the system, such an investigation may be only done if a minimum number of reports mention the same fairness issue (\intvwr{3}). After identifying the primary cause, the issue may be tested and reproduced as a baseline for mitigation (\intvwr{7}).
The incoming feedback data from model use that should go into retraining should not be used without evaluating incoming data and preprocessing (\intvwr{10}). This step should be automated (\intvwr{10}) as human involvement would be impractical with the amount of data. The evaluation of incoming data and preprocessing measures may be guided by what has been previously performed to assess data quality and to preprocess data (\intvwr{10}). For instance, sensitive attributes may be removed as it may be done in the initial preprocessing (\intvwr{10}).
Retraining should be initiated in case a fairness issue is repeatedly reported (\intvwr{3}) or regularly based on feedback data from model use (\intvwr{4}). It should also be systematically checked whether the issue is removed with retraining (\intvwr{1}). For utilizing feedback data in the most effective way possible, an active learning approach may give directions on what is needed for improvements regarding biases (\intvwr{4}).